%% file: arxiv_bytedance.tex
\documentclass[]{bytedance}

\usepackage[utf8]{inputenc}
\usepackage{eccvabbrv}
\usepackage{graphicx}
\usepackage{booktabs}
\usepackage{arydshln}
\usepackage{multirow}
\usepackage{wrapfig}
\usepackage{amsmath}
\usepackage{amssymb}
\usepackage{adjustbox}
\usepackage{float}
\usepackage[table,dvipsnames]{xcolor}

\definecolor{upcolor}{RGB}{57,182,74}
\newcommand{\up}[1]{\textcolor{upcolor}{$\uparrow$ #1}}
\newcommand{\down}[1]{\textcolor{red}{$\downarrow$ #1}}

\renewcommand{\paragraph}[1]{\vspace{1.25mm}\noindent\textbf{#1}}
\newcommand{\keywords}[1]{%
  \par\vspace{3pt}\begingroup
  \renewcommand{\and}{\unskip; }%
  \textbf{Keywords:} #1\par
  \endgroup
}

\title{%
Actor as Its Own Critic: Unifying Region Understanding and Localization via CycleGRPO
}

{\small
\author[\star]{Xin Zhang$^{1,2~}$}
\author[\star]{Haochen Wang$^{2~}$}
\author[]{Yikang Zhou$^{2~}$}
\author[]{Zhuochen Wang$^{2~}$}
\author[\dagger]{Xiangtai Li$^{2~}$}
\author[]{Robby T. Tan$^{1~}$}
}

\affiliation[]{%
{National University of Singapore$^{1}$} \quad
{ByteDance$^{2}$}
}

\abstract{
This paper introduces Actor as Its Own Critic, a unified reinforcement learning framework, Cycle Group Relative Policy Optimization (CycleGRPO), that \textit{jointly} optimizes region understanding and localization for Multimodal Large Language Models (MLLMs).
Unlike existing separate pipelines, \textit{we leverage the inherent duality between the two tasks to construct a self-evaluating reinforcement learning paradigm: ``region $\to$ text $\to$ region''}.
Specifically, a single MLLM first acts as the actor to generate region captions, then immediately transitions to a critic to ground its generated text back in the spatial domain.
Therefore, CycleGRPO requires only region inputs, \textit{e.g.}, masks or bounding boxes, entirely bypassing the need for textual ground truths.
A quality-aware token-level cycle-consistency reward is employed to assess the semantic discriminability of text captions via their physical localization accuracy.
Empirically, built upon SAMTok, our CycleGRPO framework successfully bootstraps both capabilities simultaneously.
Without any task-specific fine-tuning, the framework yields consistent performance gains across a wide range of benchmarks, including region captioning, region VQA, grounded dialogue, and referring segmentation.
Overall, CycleGRPO offers a straightforward and scalable way to advance pixel-level capabilities in MLLMs.
Code and models are released at \url{https://github.com/devinxzhang/CycleGRPO}.
}

\begin{document}
\maketitle
\begingroup
\renewcommand{\thefootnote}{}
\footnotetext{$^\star$ Equal contribution. \quad $^\dagger$ Corresponding author.}
\endgroup

\input{1_intro}

\input{3_method}

\input{4_exp}
\input{2_related_work}
\FloatBarrier
\input{5_conclusion}
\clearpage

\input{6_appendix}

\bibliographystyle{splncs04}
\bibliography{main}
\end{document}

%% file: 1_intro.tex
\section{Introduction}
\label{sec:intro}

\begin{figure}[t]
	\centering
	\includegraphics[width=1\linewidth]{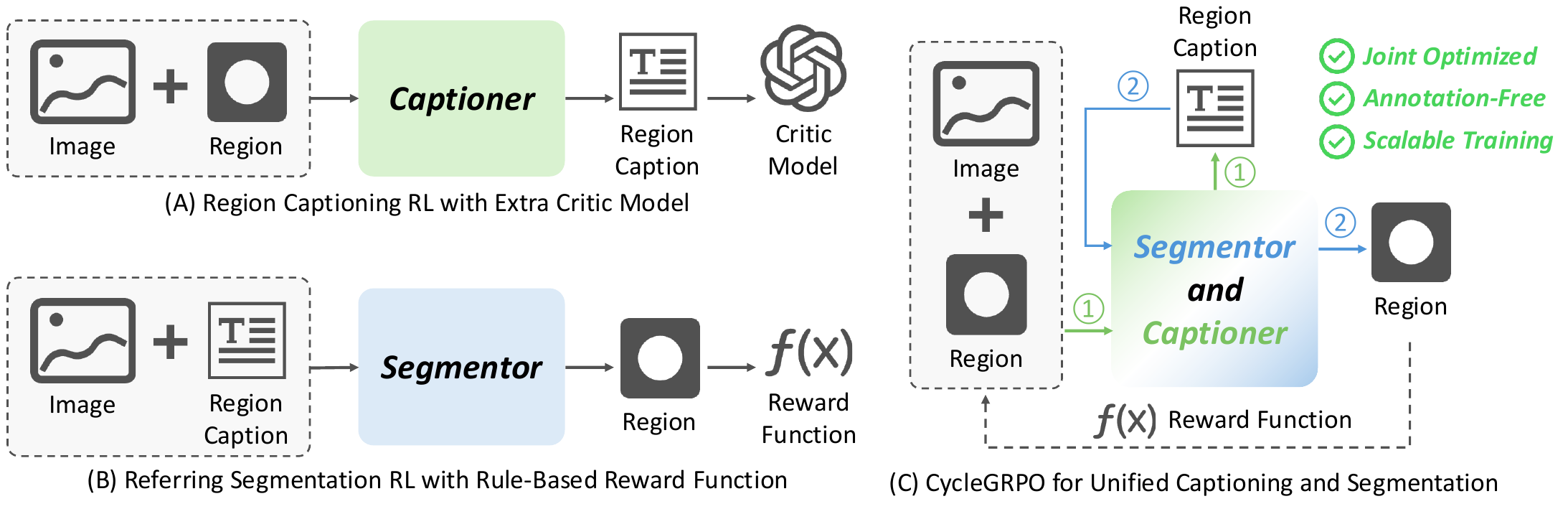}
    \caption{
    \textbf{Concept comparison of Reinforcement Learning paradigms for region-level tasks.} 
    \textbf{(A)} Captioning tasks typically rely on external LLMs to judge the text quality, which is vulnerable to reward hacking.
    \textbf{(B)} Localization tasks use objective rule-based functions (\textit{e.g.}, IoU).
    Both (A) and (B) heavily depend on \textit{unscalable} high-quality text-mask pairs.
    \textbf{(C)} We let a single model act as both the actor (generating captions) and its own critic (referring its own text back into masks). 
    This cycle consistency provides intrinsic reward signals, bypassing external judges and the need for text ground truths.}
    \label{fig:teaser}
\end{figure}

MLLMs~\cite{qwen25vl, internvl35, deepseekvl2, llava, chen2025hypospace, liu2026tele} have shown strong ability in general visual-language understanding.
Yet they still lack fine-grained capabilities required for real-world perception.
Among these, region understanding~\cite{gar, dam, osprey, guo2024regiongpt, zhou2026samtok, zhang2024omg, glamm} and referring localization~\cite{groundingsuite, gres, li2025groundingme, sa2va, zhou2026samtok, zhang2025mamba, zhang2024heap, zhang2023adaptive,du2026weatherreasonseg,lei2025hola,lei2024efficient} serve as two foundational abilities.
The former aims to describe and comprehend given regions, such as their appearance, structure, and spatial relation with others.
The latter (including grounding and segmentation) seeks to locate or segment the target region
based on free-form language queries.
Together, they build the core of fine-grained visual reasoning and are critical for various downstream applications~\cite{tan2024rle,tan2026knowing,tan2024occluded,yan2025erlora,Yan_2025_CVPR,Yan_2026_CVPR,zhang20264dpc2hatdynamicpointcloud,cao20263dot,DA-LLPose,chen2026udapose, Li_2026_CVPR, li2026bridging,yang2025erf,qian2024msnsegnet}.

Conventional approaches usually optimize these two tasks \textit{separately}, as demonstrated in Figure~\ref{fig:teaser} (A) and (B).
We observe that region understanding and referring localization are essentially \textit{dual problems}. 
Their symmetric nature naturally forms a closed reconstruction pipeline: ``region $\to$ text $\to$ region''. 
Crucially, \textit{a genuinely discriminative caption must contain sufficient unique details to seamlessly guide the model back to the exact initial mask.} 
Therefore, by exploiting this cycle consistency, the quality of text generation can be evaluated intrinsically using spatial localization accuracy.
This elegant formulation simultaneously \textit{obsoletes the need for expensive human annotations}, low-quality auto-generated data, and external LLM judges, unlocking a self-evaluating and highly scalable learning paradigm, as illustrated in Figure~\ref{fig:teaser} (C).

Motivated by this insight, we present Cycle Group Relative Policy Optimization (\textbf{CycleGRPO}), a scalable reinforcement learning (RL) pipeline that fundamentally instantiates this duality.
Built upon the discrete mask representation of SAMToK~\cite{zhou2026samtok}, which makes fine-grained RL for segmentation feasible, CycleGRPO \textit{incentivizes} the MLLM to produce discriminative captions via cycle-consistent rewards.
At its core, CycleGRPO embodies an ``Actor as Its Own Critic'' paradigm, eliminating the conventional reliance on disparate reward models or task-specific heads. 
Specifically, the model first acts as the \textit{Actor} to generate a set of candidate region descriptions. 
Then, the same model transitions into its own \textit{Critic}, which executes a rollout to localize the generated text back into the spatial domain. 
By explicitly optimizing this intrinsic cycle consistency, \textit{e.g.}, the reconstructive Intersection-over-Union (IoU) between input regions and predicted regions, CycleGRPO jointly bootstraps region understanding and localization in a fully autonomous manner.

Empirically, we comprehensively evaluate CycleGRPO across a diverse set of tasks, including region captioning, region VQA, grounded dialogue, and referring segmentation.
Our method achieves remarkable performance gains on GRES~\cite{gres}, delivering a 7.0\% increase in average gIoU compared to the SAMTok baseline. 
On GroundingSuite~\cite{groundingsuite}, CycleGRPO improves the overall grounding accuracy from 57.5\% to 67.6\%, with a significant 8.5\% boost on the challenging Part split. 
Furthermore, the framework demonstrates superior reasoning and synchronization capabilities, yielding a 5.8\% average improvement on DLC-Bench~\cite{dam}, a 0.9\% overall gain on GAR-Bench-VQA~\cite{wang2025grasp}, and a 6.5 increase in CIDEr on the GCG benchmark~\cite{glamm}. 
Notably, these consistent improvements are achieved without any task-specific fine-tuning on these datasets, underscoring that our approach provides a straightforward and scalable way to advance pixel-level capability in MLLMs by bypassing the bottleneck of textual annotations. In summary, our key contributions are:
\begin{itemize}
    \item We propose \textbf{CycleGRPO}, a unified reinforcement learning framework that fundamentally instantiates the duality between region understanding and localization, enabling a fully autonomous and scalable training pipeline.
    \item We introduce an \textit{Actor-as-Own-Critic} paradigm in which a single MLLM self-evaluates its text generation using spatial consistency rewards, thereby eliminating reliance on expensive human annotations or external LLM judges.
    \item Extensive experiments demonstrate that CycleGRPO successfully bootstraps both capabilities simultaneously, achieving highly competitive performance against heavily supervised baselines on multiple fine-grained benchmarks.
\end{itemize}

%% file: 3_method.tex
\section{Preliminary}

\noindent
\textbf{GRPO for Region Localization and Captioning.} 
RL has been widely adopted to align MLLMs with human preferences and task-specific metrics. Recently, GRPO~\cite{shao2024deepseekmath} has emerged as a highly efficient alternative to traditional Proximal Policy Optimization (PPO). 
By evaluating the relative quality within a sampled group of outputs, GRPO eliminates the need for a parameterized value network (critic), significantly reducing memory overhead while maintaining stable policy updates.

Formally, let $q$ denote a multi-modal input prompt and $\pi_\theta$ denote the MLLM actor. For a given input $q$, the policy $\pi_{\theta}$ samples a group of $G$ candidate outputs $\{y_1, y_2, \dots, y_G\}$. A reward function $R(q, y)$ evaluates these candidates to assign scalar rewards $\{r_1, r_2, \dots, r_G\}$. The advantage $A_i$ for each output $y_i$ is then computed by standardizing the rewards within the sampled group as $A_i = (r_i - \mu_r) / \sigma_r$, where $\mu_r$ and $\sigma_r$ are the mean and standard deviation of the group rewards, respectively. The model is optimized by maximizing an objective that incorporates a clipping mechanism and a Kullback-Leibler (KL) divergence penalty to prevent the policy from deviating excessively from a reference model $\pi_{\text{ref}}$:
\begin{equation}
\begin{aligned}
    \mathcal{J}_{\text{GRPO}}(\theta) = \mathbb{E}_{q, \{y_i\}_{i=1}^G} \Bigg[ \frac{1}{G} \sum_{i=1}^G \Big( & \min \big( \rho_i A_i, \text{clip}(\rho_i, 1-\epsilon, 1+\epsilon) A_i \big) \\
    & - \beta \mathbb{D}_{\text{KL}} (\pi_\theta(y_i|q) \| \pi_{\text{ref}}(y_i|q)) \Big) \Bigg],
\end{aligned}
\label{eq:grpo}
\end{equation}
where $\rho_i = \frac{\pi_\theta(y_i|q)}{\pi_{\theta_{\text{old}}}(y_i|q)}$ is the probability ratio, $\epsilon$ is the clipping threshold, and $\beta$ controls the KL penalty.

When applying this framework to fine-grained region tasks, region understanding and localization are typically treated as isolated optimization problems (as contrasted in Figure~\ref{fig:teaser} (A) and (B)):
\begin{itemize}
    \item \textbf{Region Captioning}: The input $q_{\text{cap}} = (I, M)$ consists of an image $I$ and a region mask $M$. The actor generates a text caption $y_{\text{cap}} = C$. The reward $r$ typically relies on external textual ground truths (e.g., CIDEr) or expensive LLM-as-a-judge.
    \item \textbf{Region Localization}: The input $q_{\text{loc}} = (I, C)$ consists of an image and a text description $C$. The policy generates a spatial mask $y_{\text{loc}} = \hat{M}$. The reward $r$ is often defined by the Intersection over Union (IoU) between $\hat{M}$ and the ground-truth mask $M$.
\end{itemize}

While effective, this isolated paradigm imposes a severe bottleneck: it inherently necessitates vast amounts of high-quality, human-annotated text-region pairs for supervision. To overcome this, we seek a unified representation that allows these two tasks to interact within a closed, self-supervised loop.

\noindent\textbf{Unified Mask Token Representation via SAMTok.}
\label{preliminary:samtok}
A significant challenge in applying RL to fine-grained region tasks lies in the representation gap: traditional MLLMs often separate discrete text generation from continuous mask decoding via external segmentation heads. To bridge this gap, we build our framework upon the SAMTok architecture~\cite{zhou2026samtok}.

The core of SAMTok is a discrete mask tokenizer that quantizes an arbitrary spatial mask $M$ into a compact sequence of discrete mask tokens $T_M$. By injecting these tokens into the LLM's extended vocabulary $\mathcal{V}$, a region mask is effectively translated into a format that the model natively understands. This tokenization reframes region localization as a standard autoregressive next-token prediction task, entirely eliminating the need for task-specific heads. Consequently, the MLLM can uniformly model the probability of generating either a text caption $C$ or a spatial mask $M$ within the \textit{same} probability space:
\begin{equation}
    P(y | q) = \prod_{j=1}^{L} \pi (y_j | y_{<j}, q), \quad y_j \in \mathcal{V}
\end{equation}
where $y$ can represent either a sequence of text tokens or mask tokens. This unified discrete representation serves as the structural prerequisite for our framework, allowing the same model to process both understanding and localization tasks using a single, consistent vocabulary.

\noindent\textbf{Unified Bounding Box Token Representation.}
Current advanced open-source MLLMs~\cite{qwen3vl, qwen25vl, internvl35} already support unified token representation for text and bounding boxes.
They serve as box coordinates, treated as regular text tokens, eliminating the need for dedicated detection heads.
To demonstrate the generalization capabilities of our CycleGRPO, we also conduct experiments on the Qwen3-VL~\cite{qwen3vl} benchmark for region understanding and grounding tasks.

\section{CycleGRPO: Actor as Its Own Critic}
\label{sec:method}

\begin{figure}[t]
    \centering\small
    \includegraphics[width=1.0\linewidth]{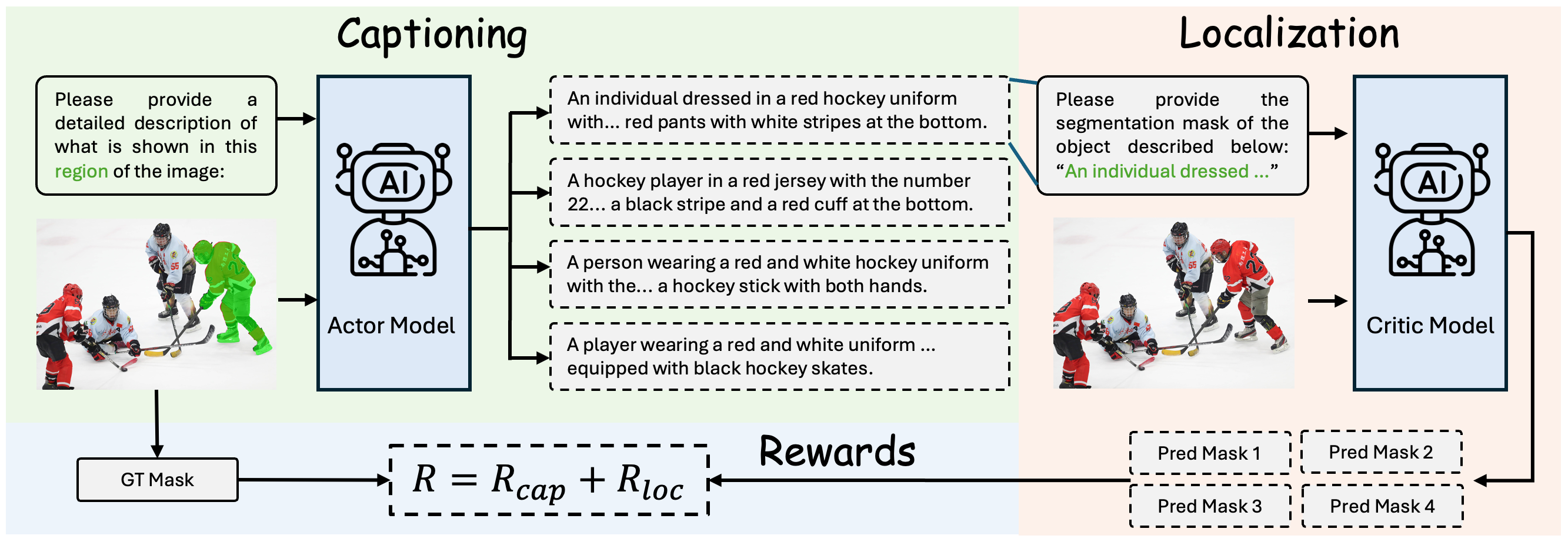}
    \caption{\small \textbf{Overview of the CycleGRPO Pipeline.} Our framework leverages a single MLLM to instantiate a self-evaluating closed loop. \textbf{Left (Phase 1: Region Captioning):} Acting as the \textit{Actor}, the model takes an image and a target region $M$ to generate a group of candidate text descriptions $\{C_1, \dots, C_G\}$. \textbf{Right (Phase 2: Localization Rollout):} Symmetrically acting as its own \textit{Critic}, the exact same model receives each candidate and attempts to ground it back into the spatial domain, yielding multiple predicted reconstructions $\{\hat{M}_{i,k}\}$. \textbf{Bottom:} The semantic quality of the captions is materialized through the spatial consistency between the predicted and original regions. The framework computes a joint cyclic reward based on this alignment, optimizing both understanding and grounding capabilities simultaneously without requiring external text annotations.}
    \label{fig:pipeline}
\end{figure}

In this section, we use region \textit{understanding} and \textit{segmentation} with SAMTok~\cite{zhou2026samtok} by default to illustrate our framework, since unifying region understanding and grounding with Qwen3-VL~\cite{qwen3vl} is nearly the same.

\noindent\textbf{Duality of Region Captioning and Localization.} 
Fundamentally, region captioning (understanding) and region localization (segmentation/grounding) can be formulated as dual problems that perform inverse transformations between visual-spatial and textual-semantic domains. 
Given an image $I$, region captioning aims to map a spatial region, defined by a binary mask $M$, to a discriminative text description $C$. Conversely, region localization grounds a description $C$ back into the physical space, predicting the corresponding mask $\hat{M}$.

Building upon the unified vocabulary enabled by SAMTok (Sec.~\ref{preliminary:samtok}), we can reformulate these two mappings as conditional probability distributions modeled by a \textit{single} policy $\pi_\theta$. The duality implies that these tasks are inherently symmetric and mutually conditional: \textit{a high-quality caption $C$ should contain sufficient discriminative details to enable the precise reconstruction of its original region.}

Formally, if $\pi_\theta$ generates an optimal caption $C^*$ via the understanding mapping $C^* \sim \pi_\theta(\cdot | I, M)$, the inverse mapping $\pi_\theta(\cdot | I, C^*)$ should ideally recover the initial spatial mask. Formally, the dual-task consistency aims to \textit{minimize the expected reconstruction error across the cycle}:
\begin{equation}
    \mathcal{L}_{\text{cycle}}(\theta) = \mathbb{E}_{C \sim \pi_\theta(\cdot|I, M)} \left[ \mathcal{D} \big( M, \pi_\theta(\cdot|I, C) \big) \right],
\label{eq:circle}
\end{equation}
where $\mathcal{D}(\cdot, \cdot)$ denotes a distance metric (e.g., IoU) in the spatial representation space. The model first acts as an actor to generate a semantic bridge $C$, and subsequently validates it through spatial reconstruction as a critic.

Unlike traditional RL pipelines that isolate these processes, this dual perspective reveals that the model's localization capability can serve as an \textit{intrinsic}, physically grounded metric for evaluating its own semantic understanding. This insight directly motivates our \textbf{Actor-as-Own-Critic} paradigm, as illustrated in Figure~\ref{fig:teaser}(C). 
In this closed loop, the model first acts as the \textit{actor} to generate candidate captions, then immediately transitions to its own \textit{critic} to perform inverse localization. By natively minimizing the cycle discrepancy without external textual ground truth, the framework directly ties the optimization of semantic discriminability to physical grounding accuracy.

\noindent\textbf{The CycleGRPO Framework.} 
Leveraging a unified representation space, we propose \textbf{CycleGRPO}.
As illustrated in Figure~\ref{fig:pipeline}, the MLLM $\pi_\theta$ dynamically transitions between a region captioner and a region localizer, forming a self-evaluating closed loop. Given an image $I$ and a target region mask $M$, the framework executes this cycle via a two-stage rollout process:

\textbf{Phase 1: Captioning Rollout (Region Understanding).} 
In the first phase, the model performs semantic exploration. We define a \textit{captioning prompt} $q_{\text{cap}} = (I, Q_{\text{cap}}(M))$, where $Q_{\text{cap}}(\cdot)$ is a task-specific instruction template that embeds the target region. The policy $\pi_\theta$ then autoregressively samples a group of $G$ candidate captions:
\begin{equation}
    \{C_1, C_2, \dots, C_G\} \sim \pi_\theta(\cdot \mid q_{\text{cap}}).
\end{equation}
In the absence of textual ground truths or external judges, these candidates represent a diverse set of semantic hypotheses regarding the visual region.

\textbf{Phase 2: Localization Rollout (Region Localization).} 
To evaluate these hypotheses, the model transitions into the role of a localizer to close the cycle. For each candidate caption $C_i$, we construct a \textit{localizing prompt} $q_{\text{loc}}^{(i)} = (I, Q_{\text{loc}}(C_i))$ to query the model for spatial reconstruction.

A key challenge here is the inherent stochasticity of autoregressive decoding: a single greedy or sampled prediction $\hat{M}_i$ may not fully reflect the discriminative potential of $C_i$. To obtain a more robust proxy for semantic quality, we perform $K$ independent localizing rollouts for each candidate $C_i$, yielding a set of predicted spatial masks:
\begin{equation}
    \{\hat{M}_{i,1}, \dots, \hat{M}_{i,K}\} \sim \pi_\theta(\cdot \mid q_{\text{loc}}^{(i)}), \quad \forall i \in \{1, \dots, G\}.
\end{equation}
By aggregating the spatial accuracy across these $K$ rollouts, the framework obtains a robust estimate of the caption's quality, effectively suppressing the noise inherent in autoregressive sampling.

Through this dual-stage process, the abstract semantic quality of $C_i$ is materialized into concrete spatial reconstructions $\{\hat{M}_{i,k}\}_{k=1}^K$. If the generated caption $C_i$ is accurate and unambiguous, the grounding rollout should consistently recover the initial spatial representation $M$. This materialization allows us to bypass the need for external textual supervision, as the spatial consistency between $M$ and $\hat{M}$ now serves as the foundational signal for the reward mechanism.

\noindent\textbf{The Cycle Reward. }
The CycleGRPO framework optimizes the model by converting the spatial consistency between the initial region $M$ and its cyclic reconstructions $\{\hat{M}_{i,k}\}$ into training signals for both tasks.
For each grounding rollout $\hat{M}_{i,k}$, we compute an alignment score $s_{i,k}$ that measures its spatial consistency with the reference mask $M$. Following standard practice in visual grounding, we define this score as the Intersection over Union (IoU) between the masks, denoted as $s_{i,k} = \text{IoU}(M, \hat{M}_{i,k})$. Using IoU provides a continuous and physically grounded signal that accurately reflects the quality of the reconstruction.
Specifically, this alignment score is then used to supervise both the captioning and grounding phases:
\begin{itemize}
    \item \textbf{Captioning Reward ($R_i^{\text{cap}}$):} The quality of the $i$-th candidate caption $C_i$ is evaluated by the average IoU of its $K$ corresponding grounding paths: $R_i^{\text{cap}} = \frac{1}{K} \sum_{k=1}^K s_{i,k}$. A low $R_i^{\text{cap}}$ indicates that the caption is either hallucinated or too vague to facilitate precise reconstruction.
    \item \textbf{Localization Reward ($R_{i,k}^{\text{loc}}$):} For the grounding phase, we weight each trajectory by its parent caption's reward: $R_{i,k}^{\text{loc}} = R_i^{\text{cap}} \cdot s_{i,k}$. This weighting ensures that gradients from high-quality captions are amplified, while those from low-quality or hallucinated text are suppressed.
\end{itemize}

The final training reward for a complete spatial-semantic cycle is defined as the sum of the two components: $R^{\text{total}} = R^{\text{cap}} + R^{\text{loc}}$. 
By maximizing this joint reward, the model is incentivized to generate captions that are not only linguistically fluent but also spatially discriminative, while simultaneously refining its grounding precision. 
This total reward is subsequently standardized within each sampling group to compute the advantages $A^{\text{cap}}$ and $A^{\text{loc}}$ as required by the GRPO objective (Eq.~\ref{eq:grpo}).

%% file: 4_exp.tex
\section{Experiment}
\label{sec:exp}

\noindent
\textbf{Implementation Details.}
We implement the CycleGRPO framework using the \texttt{verl} RL library. Our base MLLM is SAMTok~\cite{zhou2026samtok}, built upon the Qwen3-VL-4B~\cite{qwen3vl} backbone. Note that we employ the pre-trained version of SAMTok without any further supervised fine-tuning (SFT) or RL on specific downstream datasets. During training, we freeze the vision encoder and fine-tune the projection layer and the LLM parameters. For the training corpus, we sample 20k images with corresponding object masks from DenseWorld~\cite{li2025denseworld}, supplemented by 1k no-target expressions from GRES~\cite{gres} to enhance discriminative robustness.
The model is trained for one epoch with a total batch size of 128. Following the group-sampling strategy of GRPO, we set both the caption group size $G$ and the grounding rollout count $K$ to 6, yielding $G \times K = 36$ spatial trajectories per image to estimate the cycle reward. We use the AdamW optimizer with a learning rate of $1 \times 10^{-6}$ and a weight decay of $1 \times 10^{-2}$. 

\noindent\textbf{Dataset and Benchmarks.}
To evaluate the effectiveness of CycleGRPO, we conduct experiments on two representative task categories: region captioning and region localization. 
Note that we do not perform task-specific fine-tuning on any of the evaluation benchmarks. 
Instead, a single unified model is evaluated directly on all benchmarks for both tasks, demonstrating generalization and robust dual-task alignment.
For region captioning, we use DLC-Bench~\cite{dam} and GAR-Bench-VQA~\cite{wang2025grasp} to assess discriminative language capability and descriptive richness. For region localization, we evaluate spatial grounding performance on GCG~\cite{glamm}, GRES~\cite{gres}, and the GroundingSuite~\cite{groundingsuite} (covering \textit{Stuff}, \textit{Part}, and \textit{Multi-object} scenarios).
GRES is subsequently used to evaluate the model's target-rejection capability.

\noindent
\textbf{Comparison Baselines.}
We compare our CycleGRPO against a comprehensive suite of state-of-the-art models, including: (1) General MLLMs, such as the private GPT-4o~\cite{gpt4o}, o3~\cite{o3}, and Gemini 2.5 Pro~\cite{gemini-2.5-pro}, alongside the open-source Qwen2.5-VL~\cite{qwen25vl} and InternVL3 series~\cite{internvl3}; (2) Region-level and Segmentation MLLMs, encompassing mask-based models like LISA~\cite{lisa}, GLaMM~\cite{glamm}, Sa2VA~\cite{sa2va}, VP-SPHINX~\cite{lin2024VP-SPHINX}, EVF-SAM~\cite{zhang2024evf-sam}, SAMTok, and ARGenSeg~\cite{argenseg}, as well as specialized referring segmentation models such as SAM4MLLM~\cite{chen2024sam4mllm}, MLLMSeg~\cite{wang2025MLLMSeg}, and HiMTok~\cite{himtok}; and (3) Task-specific Baselines, including DAM~\cite{dam} for region understanding and InstructSeg~\cite{wei2025instructseg} for grounding tasks.

\begin{table}[!tbp]
\begin{minipage}[t]{0.4\linewidth}
\vspace{0pt}
\centering\small
\setlength{\tabcolsep}{3pt}
\caption{
Results on the mask-to-text task (DLC-Bench~\cite{dam}).
%
%
%
$^{\dag}$ indicates trained on \textit{in-domain} DAM-1.5M~\cite{dam} with respect to DLC-Bench~\cite{dam}.
}
\label{tab:dlc_bench}
\resizebox{1\columnwidth}{!}{
\begin{tabular}{l cccc}
\toprule
\textbf{Method} & \textbf{Size} & \textbf{Pos.} & \textbf{Neg.} & \textbf{Avg.} \\
\midrule
GPT-4o~\cite{gpt4o} & -- & 43.4 & 79.6 & 61.5 \\
Gemini 2.5 Pro~\cite{gemini} & -- & 36.5 & 75.2 & 55.8 \\
DAM$^{\dag}$~\cite{dam} & 3B & 52.3 & 82.2 & 67.3 \\
\midrule
SAMTok~\cite{zhou2026samtok} & 4B & 43.5 & 80.4 & 61.9 \\
\ + GRPO & 4B & 45.0 & 81.8 & 63.4 \\
\ + CycleGRPO & 4B & 51.2 & 84.2 & 67.7 \\
$\Delta$ \textit{v.s.} SAMTok~\cite{zhou2026samtok} & & \up{7.7} & \up{3.8} & \up{5.8} \\
$\Delta$ \textit{v.s.} GRPO & & \up{6.2} & \up{2.4} & \up{4.3} \\
\bottomrule
\end{tabular}}
\end{minipage}
\hfill
\begin{minipage}[t]{0.58\linewidth}
\vspace{0pt}
\centering\small
\caption{
Results on text-to-mask task (GroundingSuite~\cite{groundingsuite}). 
We report the gIoU to evaluate the model's zero-shot generalization capability across diverse spatial grounding scenarios.
}
\label{tab:groundingsuite}
\setlength{\tabcolsep}{3pt}
\resizebox{1\columnwidth}{!}{
\begin{tabular}{lc ccccc}
\toprule
\textbf{Method} & \textbf{Size} & \textbf{Stuff} & \textbf{Part} & \textbf{Multi} & \textbf{Single} & \textbf{All} \\
\midrule
LISA~\cite{lisa} & 7B & 85.2 & 21.2 & 71.5 & 42.8 & 57.6 \\
GLaMM~\cite{glamm} & 7B & 86.9 & 16.5 & 70.4 & 42.1 & 57.2 \\
EVF-SAM~\cite{zhang2024evf-sam} & 1B & 85.1 & 23.1 & 72.1 & 54.5 & 62.6 \\
InstructSeg~\cite{wei2025instructseg} & 3B & 56.2 & 24.2 & 66.8 & 51.3 & 52.5 \\
\midrule
SAMTok~\cite{zhou2026samtok} & 4B & 80.9 & 12.4 & 62.0 & 52.9 & 57.5 \\
\ + GRPO & 4B & 89.3 & 16.5 & 71.6 & 54.1 & 62.7 \\
\ + CycleGRPO & 4B & 90.7 & 20.9 & 76.3 & 61.6 & 67.6 \\
$\Delta$ \textit{v.s.} SAMTok~\cite{zhou2026samtok} & & \up{9.8} & \up{8.5} & \up{14.3} & \up{8.7} & \up{10.1} \\
$\Delta$ \textit{v.s.} GRPO & & \up{1.4} & \up{4.4} & \up{4.7} & \up{7.5} & \up{4.9} \\
\bottomrule
\end{tabular}}
\end{minipage}
\end{table}

\subsection{Main Results}

\begin{figure}[!htbp]
    \centering\small
    \includegraphics[width=1.0\linewidth]{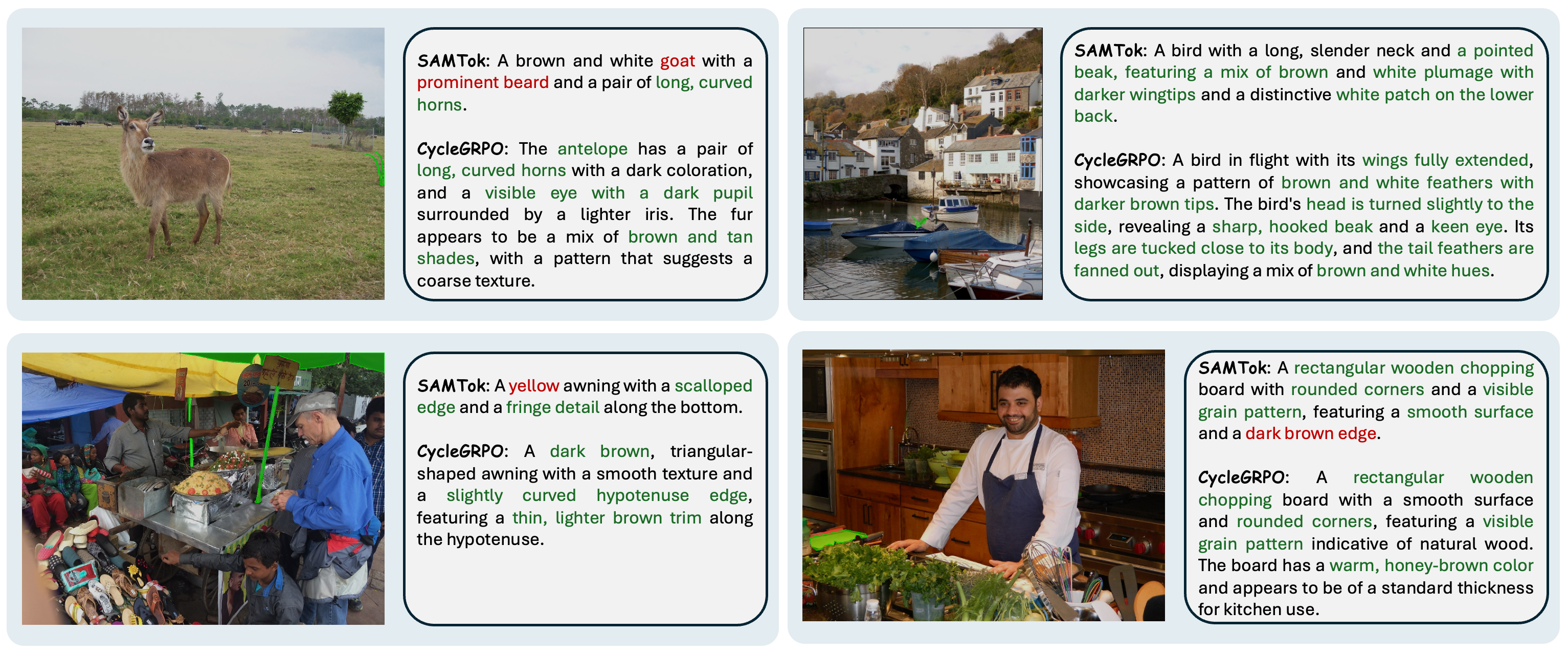}
    \caption{\small \textbf{Qualitative comparison of region captioning on DLC-Bench~\cite{dam}.} We compare the descriptions generated by SAMTok and CycleGRPO for the regions \textbf{highlighted in green}. In the text, \textcolor{green}{green} indicates accurate attributes and \textcolor{red}{red} denotes errors. Compared to SAMTok~\cite{zhou2026samtok}, CycleGRPO produces more \textbf{factually accurate} descriptions (e.g., correctly identifying the ``antelope'' instead of a ``goat'') and provides significantly \textbf{richer details} regarding texture, shape, and color. These cases demonstrate that our cycle-consistency reward effectively encourages the model to distill comprehensive semantic information from visual regions.}
    \label{fig:cap}
\end{figure}

\noindent\textbf{Region Captioning.} 
Table~\ref{tab:dlc_bench} summarizes the performance on DLC-Bench, reported under the ``Llama3.1-8B without images'' evaluation mode. CycleGRPO boosts the base SAMTok model's average score from 61.9 to 67.7, outperforming closed-source frontiers like GPT-4o (61.5) and Gemini 2.5 Pro (55.8). Notably, it matches DAM (67.3) despite DAM being trained on 1.5M in-domain region-caption pairs that our model never encountered, suggesting that the self-evaluating loop in CycleGRPO can substitute for massive human annotations by bootstrapping from the model's internal spatial-semantic duality. Compared to standard GRPO (63.4), our cycle-consistent approach yields an additional 4.3 gain, indicating that intrinsic spatial grounding provides a more precise supervisory signal than generic RL. These gains are qualitatively reflected in Fig.~\ref{fig:cap}, where CycleGRPO generates more discriminative, factually grounded descriptions and reduces semantic hallucinations by anchoring text to precise spatial reconstructions.

\begin{table*}[!tbp]
\centering\small
\caption{Comparison on GAR-Bench-VQA~\cite{gar}. 
This benchmark consists of multiple-choice questions designed to evaluate the model's fine-grained perception and reasoning capabilities. 
We report the accuracy (\%) across various categories, where $^{\dag}$ denotes models evaluated in thinking mode. 
Our method demonstrates superior zero-shot performance, particularly in complex perception and non-entity reasoning, even without task-specific tuning or human-annotated VQA data.
}
\label{tab:gar_bench}
\setlength{\tabcolsep}{3pt}
\resizebox{1\columnwidth}{!}{
\begin{tabular}{l c cccccccc}
\toprule
\multirow{2}{*}{Method} & \multirow{2}{*}{Size} & \multirow{2}{*}{Overall} 
& \multicolumn{4}{c}{Perception} 
& \multicolumn{3}{c}{Reasoning} \\
\cmidrule(lr){4-7} \cmidrule(lr){8-10}
& & & Color & Shape & Texture & Material
& Position & Non-Entity & Relation \\
\midrule
GPT-4o~\cite{gpt4o} & - & 53.5 & 34.8 & 65.3 & 48.3 & 52.8 & 57.8 & 60.2 & 61.4 \\
o3$^{\dagger}$~\cite{o3} & - & 61.3 & 58.0 & 70.3 & 55.2 & 63.9 & 54.7 & 49.2 & 71.3 \\
Gemini-2.5-Pro$^{\dagger}$~\cite{gemini-2.5-pro} & - & 64.2 & 62.3 & 68.8 & 58.6 & 66.7 & 64.1 & 64.9 & 70.3 \\
\midrule
Qwen2.5-VL~\cite{qwen25vl} & 3B & 34.4 & 29.0 & 25.0 & 34.5 & 30.6 & 43.8 & 26.2 & 44.6 \\
Qwen2.5-VL~\cite{qwen25vl} & 7B & 41.7 & 39.1 & 40.6 & 44.8 & 27.8 & 59.4 & 36.1 & 40.6 \\
Qwen2.5-VL~\cite{qwen25vl} & 32B & 50.9 & 46.4 & 53.1 & 41.4 & 30.6 & 71.9 & 36.1 & 58.4 \\
Qwen2.5-VL~\cite{qwen25vl} & 72B & 52.8 & 46.4 & 50.0 & 65.5 & 33.3 & 68.8 & 44.3 & 57.4 \\
InternVL3~\cite{internvl3} & 2B & 35.1 & 30.4 & 21.9 & 48.3 & 38.9 & 48.4 & 26.2 & 38.6 \\
InternVL3~\cite{internvl3} & 8B & 38.9 & 36.2 & 37.5 & 58.6 & 41.7 & 51.6 & 27.9 & 33.6 \\
InternVL3~\cite{internvl3} & 32B & 46.5 & 39.1 & 40.6 & 51.7 & 55.6 & 60.9 & 36.1 & 47.5 \\
InternVL3~\cite{internvl3} & 78B & 50.5 & 44.9 & 54.7 & 58.6 & 61.1 & 53.1 & 47.5 & 45.5 \\
\midrule
Sa2VA~\cite{sa2va} & 8B & 34.3 & 39.1 & 45.3 & 29.6 & 30.6 & 54.7 & 21.3 & 21.8 \\
VP-SPHINX~\cite{lin2024VP-SPHINX} & 13B & 37.5 & 33.3 & 25.0 & 44.8 & 38.9 & 60.9 & 34.3 & 32.7 \\
DAM~\cite{dam} & 3B & 38.2 & 55.1 & 39.1 & 41.4 & 36.1 & 31.3 & 36.1 & 31.7 \\
GAR~\cite{gar} & 1B & 50.6 & 55.1 & 46.9 & 69.0 & 47.2 & 21.9 & 62.3 & 56.4 \\
GAR~\cite{gar} & 8B & 59.9 & 59.4 & 54.7 & 75.9 & 52.8 & 48.4 & 60.7 & 68.3 \\
\midrule
SAMTok~\cite{zhou2026samtok} & 4B & 64.2 & 58.0 & 48.4 & 48.3 & 58.3 & 76.6 & 54.1 & 83.2 \\
\ + GRPO & 4B & 63.9 & 58.0 & 46.9 & 44.8 & 58.3 & 76.6 & 54.1 & 84.2\\
\ + CycleGRPO & 4B & 65.1 & 62.3 & 50.0 & 48.3 & 61.1 & 73.4 & 57.4 & 82.2 \\
$\Delta$ \textit{v.s.} SAMTok~\cite{zhou2026samtok} & & \up{0.9} & \up{4.3} & \up{1.6} & \textcolor{gray}{-- 0.0} & \up{2.8} & \down{3.2} & \up{3.3} & \down{1.0} \\
$\Delta$ \textit{v.s.} GRPO & & \up{1.2} & \up{4.3} & \up{3.1} & \up{3.5} & \up{2.8} & \down{3.2} & \up{3.3} & \down{2.0} \\
\bottomrule
\end{tabular}}
\end{table*}

\noindent\textbf{Region VQA.}
As shown in Table~\ref{tab:gar_bench}, CycleGRPO elevates SAMTok's overall score to 65.1\%, surpassing frontier closed-source models such as Gemini-2.5-Pro (64.2\%) and o3 (61.3\%). The improvement is particularly pronounced in fine-grained \textit{Perception}, where our model delivers consistent gains in categories such as Color (+4.3\%), Shape (+1.6\%), and Material (+2.8\%). Furthermore, in complex \textit{Reasoning}, CycleGRPO achieves a significant 3.3\% boost in the Non-Entity split compared to the SAMTok baseline. 
The slight drops on the Position (-3.2\%) and Relation (-1.0\%) splits stem from the nature of our reward: mask reconstruction depends more on attribute descriptions (e.g., color, shape) than on spatial relations, steering the model to prioritize the former.
Compared to naive GRPO, which shows negligible or even negative impact on this benchmark (63.9\% vs. 64.2\%), our cycle-consistency reward provides a more effective training signal by anchoring textual descriptions to physical spatial reconstruction. These results demonstrate that CycleGRPO successfully distills superior visual-reasoning capabilities, achieving better factual grounding without requiring any task-specific tuning or human-annotated VQA data.

\noindent\textbf{Interleaved Text-mask Generation.} 
The GCG benchmark (Table~\ref{tab:gcg}) jointly evaluates understanding and grounding by requiring simultaneous generation of descriptive text and precise masks. CycleGRPO consistently outperforms the SAMTok baseline, with a significant leap in captioning quality (+6.5 CIDEr, +1.1 METEOR on Val) alongside improved grounding (+3.0\% Recall, +1.2\% AP$_{50}$). Compared to standard GRPO (+3.8 CIDEr), our cycle-consistency objective provides a more effective signal for semantic-spatial synchronization. Notably, despite its 4B scale, our model surpasses larger counterparts like GLaMM (7B) and Sa2VA (8B) in both descriptive richness and localization accuracy.

\begin{table}[!tbp]
\centering
\caption{
Results on interleaved text-mask generation task (GCG)~\cite{glamm}.
This benchmark provides a holistic evaluation of the model's \textbf{joint understanding and grounding} capabilities by requiring the simultaneous generation of descriptive text and precise segmentation masks. 
We report linguistic metrics (METEOR, CIDEr) alongside spatial metrics (AP$_{50}$, mIoU, Recall) to assess the semantic-spatial synchronization. 
CycleGRPO consistently improves both text quality and localization accuracy.
}
\label{tab:gcg}
\setlength{\tabcolsep}{3pt}
\resizebox{1\columnwidth}{!}{%
\begin{tabular}{l c ccccc ccccc}
\toprule
\multirow{2}{*}{\textbf{Method}} & \multirow{2}{*}{\textbf{Size}} 
& \multicolumn{5}{c}{\textbf{Val}} 
& \multicolumn{5}{c}{\textbf{Test}} \\
\cmidrule(lr){3-7}
\cmidrule(lr){8-12}

& & METEOR & CIDEr & AP50 & mIoU & Recall
  & METEOR & CIDEr & AP50 & mIoU & Recall \\
\midrule

LISA~\cite{lisa} & 7B 
& 13.0 & 33.9 & 25.2 & 62.0 & 36.3
& 12.9 & 32.2 & 24.8 & 61.7 & 35.5 \\

GLaMM~\cite{glamm} & 7B
& 16.2 & 47.2 & 30.8 & 66.3 & 41.8
& 15.8 & 43.5 & 29.2 & 65.6 & 40.8 \\

OMG-LLaVA~\cite{zhang2024omg}  & 7B
& 14.9 & 41.2 & 29.9 & 65.6 & --
& 14.5 & 38.5 & 28.6 & 64.7 & -- \\

Sa2VA~\cite{sa2va}  & 8B
& 16.4 & 49.5 & 33.2 & 67.7 & 45.1
& 16.2 & 49.0 & 32.2 & 66.8 & 44.5 \\

\midrule

SAMTok~\cite{zhou2026samtok} & 4B
& 16.1 & 48.2 & 34.7 & 69.4 & 46.6
& 16.4 & 51.4 & 34.4 & 68.4 & 48.3 \\
\ + GRPO & 4B & 16.4 & 50.9 & 35.8 & 69.6 & 47.5 & 16.4 & 52.3 & 35.3 & 68.9 & 47.9 \\
\ + CycleGRPO & 4B 
& 17.2 & 54.7 & 35.9 & 69.6 & 49.6 
& 17.1 & 54.0 & 35.2 & 68.6 & 49.7 \\
$\Delta$ \textit{v.s.} SAMTok~\cite{zhou2026samtok} & & \up{1.1} & \up{6.5} & \up{1.2} & \up{0.2} & \up{3.0} & \up{0.7} & \up{2.6} & \up{0.8} & \up{0.2} & \up{1.4} \\
$\Delta$ \textit{v.s.} GRPO & & \up{0.8} & \up{3.8} & \up{0.1} & \textcolor{gray}{- 0.0} & \up{2.1} & \up{0.7} & \up{1.7} & \down{0.1} & \down{0.3} & \up{1.8} \\
\bottomrule
\end{tabular}}
\end{table}

\begin{figure}[!htbp]
    \centering\small
    \includegraphics[width=1.0\linewidth]{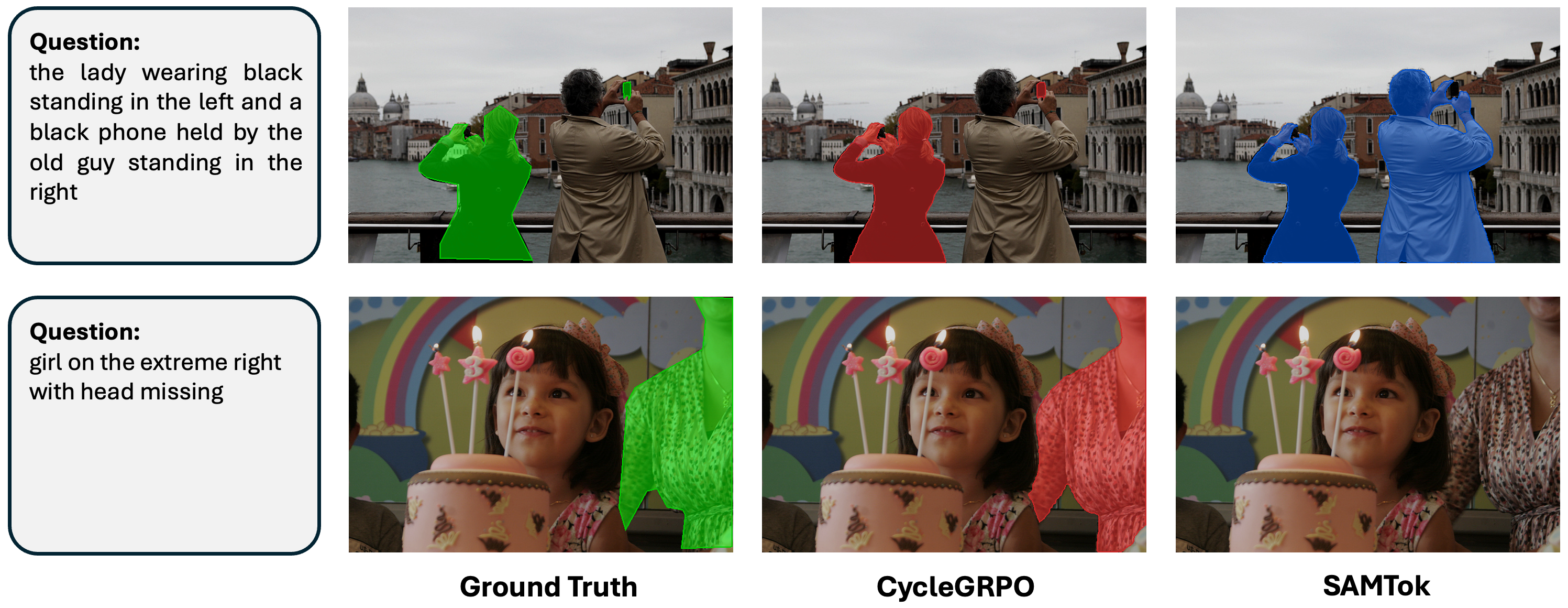}
    \caption{\small \textbf{Qualitative results on GRES~\cite{gres}.} We compare CycleGRPO with SAMTok~\cite{zhou2026samtok} and the ground truth. These examples require advanced spatial reasoning to distinguish targets in complex scenes. In the first row, CycleGRPO accurately localizes multiple disjoint entities mentioned in the query, whereas SAMTok incorrectly includes the man's coat. In the second row, our model correctly identifies the target despite occlusions, while the baseline fails to provide a valid segmentation.}
    \label{fig:gres}
\end{figure}

\begin{figure}[!htbp]
    \centering\small
    \includegraphics[width=1.0\linewidth]{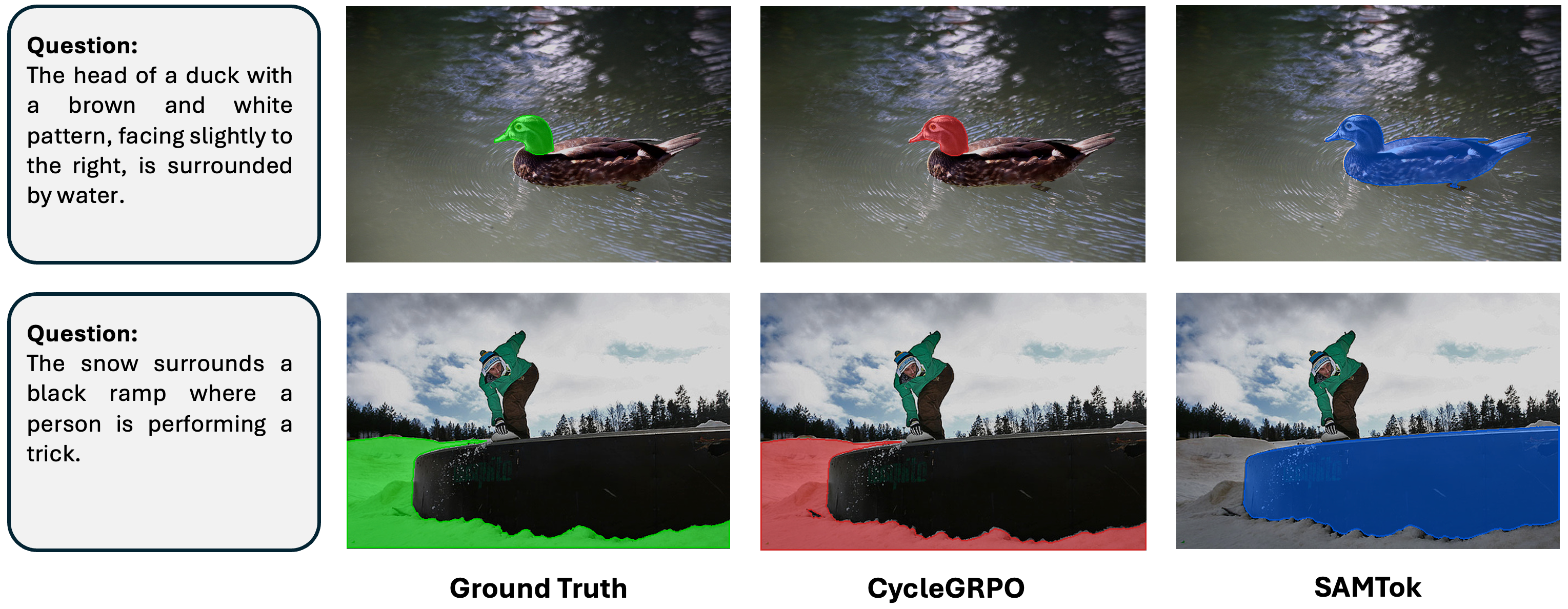}
    \caption{\small \textbf{Qualitative results on GroundingSuite~\cite{groundingsuite}.} We compare our CycleGRPO with SAMTok~\cite{zhou2026samtok} and the ground truth. The first row demonstrates \textbf{part-level segmentation} (e.g., the duck's head), which requires fine-grained understanding of object components. The second row illustrates \textbf{stuff-class segmentation} (e.g., the snow), highlighting the model's capability for context-aware localization in amorphous regions. CycleGRPO consistently produces masks that are more precisely aligned with the textual queries.}
    \label{fig:groundingsuite}
\end{figure}

\noindent\textbf{Text-to-Mask.} 
We evaluate spatial grounding on the GRES and GroundingSuite benchmarks. As shown in Table~\ref{tab:gres}, CycleGRPO achieves a dramatic leap on GRES: the model's ability to reject non-existent targets (Avg. N-acc) soars from 58.7\% to 92.1\%, indicating that our cyclic objective curbs hallucinations by penalizing captions that lead to failed spatial reconstructions. It also delivers a 7.0\% gain in average gIoU over SAMTok, outperforming dedicated segmentation models such as LISA (62.2\%) and ARGenSeg (73.6\%). On GroundingSuite (Table~\ref{tab:groundingsuite}), CycleGRPO lifts overall accuracy from 57.5\% to 67.6\%, with substantial gains on Multi-object (+14.3\%) and Single-object (+8.7\%) tasks. These improvements are visually evident in Fig.~\ref{fig:gres} and Fig.~\ref{fig:groundingsuite}.

\begin{table*}[!tbp]
\centering
\small
\setlength{\tabcolsep}{2pt}
\caption{
Results on text-to-mask task (GRES)~\cite{gres}. 
This benchmark evaluates Referring Expression Segmentation (RES) and target rejection capability. We report gIoU, cIoU, and N-acc (\%).
}
\label{tab:gres}
\resizebox{1\columnwidth}{!}{%
\begin{tabular}{l c ccc ccc ccc ccc}
\toprule

\multirow{2}{*}{\textbf{Method}} &
\multirow{2}{*}{\textbf{Size}} &
\multicolumn{3}{c}{\textbf{Val}} &
\multicolumn{3}{c}{\textbf{Test A}} &
\multicolumn{3}{c}{\textbf{Test B}} &
\multicolumn{3}{c}{\textbf{Avg.}} \\

\cmidrule(lr){3-5}
\cmidrule(lr){6-8}
\cmidrule(lr){9-11}
\cmidrule(lr){12-14}

& & gIoU & cIoU & N-acc
  & gIoU & cIoU & N-acc
  & gIoU & cIoU & N-acc
  & gIoU & cIoU & N-acc \\

\midrule

LISA~\cite{lisa}  & 7B
& 61.6 & 61.8 & 54.7
& 66.3 & 68.5 & 50.0
& 58.8 & 60.6 & 51.9
& 62.2 & 63.6 & 52.2 \\

SAM4MLLM~\cite{chen2024sam4mllm}  & 8B
& 71.9 & 67.8 & 66.1
& 74.2 & 72.2 & 63.9
& 65.3 & 63.4 & 60.0
& 70.5 & 67.8 & 63.3 \\

MLLMSeg~\cite{wang2025MLLMSeg}  & 8B
& 75.1 & 71.6 & 73.2
& 77.0 & 76.9 & 72.4
& 69.7 & 68.5 & 65.5
& 73.9 & 72.3 & 70.4 \\

HiMTok~\cite{himtok}  & 8B
& 72.1 & 70.4 & --
& 73.5 & 74.9 & --
& 71.7 & 72.0 & --
& 72.4 & 72.4 & -- \\

ARGenSeg~\cite{argenseg}  & 8B
& 74.7 & 72.2 & --
& 73.7 & 73.6 & --
& 72.4 & 70.4 & --
& 73.6 & 72.1 & -- \\

\midrule

SAMTok~\cite{zhou2026samtok} & 4B
& 71.3 & 69.2 & 61.4
& 75.3 & 75.4 & 59.0
& 66.9 & 66.0 & 55.6
& 71.2 & 70.2 & 58.7 \\

\ + GRPO & 4B 
& 75.7 & 68.3 & 98.8
& 70.2 & 69.8 & 98.1 
& 66.4 & 64.0 & 96.8
& 70.8 & 67.4 & 97.9 \\

\ + CycleGRPO & 4B 
& 81.8 & 74.6 & 94.2
& 79.9 & 77.8 & 93.1 
& 73.0 & 70.0 & 89.0
& 78.2 & 74.1 & 92.1 \\
$\Delta$ \textit{v.s.} SAMTok~\cite{zhou2026samtok} & & \up{10.5} & \up{5.4} & \up{32.8} & \up{4.6} & \up{2.4} & \up{34.1} & \up{6.1} & \up{4.0} & \up{33.4} & \up{7.0} & \up{3.9} & \up{33.4} \\
$\Delta$ \textit{v.s.} GRPO & & \up{6.1} & \up{6.3} & \down{4.6} & \up{9.7} & \up{8.0} & \down{5.0} & \up{6.6} & \up{6.0} & \down{7.8} & \up{7.4} & \up{6.7} & \down{5.8} \\

\bottomrule
\end{tabular}}
\end{table*}

\subsection{Ablation and Analysis}

\begin{table}[!tbp]
\centering
\caption{Ablation on per-task GRPO baselines on SAMTok pretrained on Qwen3-VL-4B. Each single-direction variant (Captioning-only or Localization-only GRPO) improves only its own task, while CycleGRPO improves both and surpasses each baseline on its home metric.}
\label{tab:per_task_grpo}
\resizebox{0.75\columnwidth}{!}{%
\begin{tabular}{lccc|ccccc}
\toprule
\multirow{2}{*}{\textbf{Method}} & \multicolumn{3}{c|}{\textbf{DLC-Bench}} & \multicolumn{5}{c}{\textbf{GroundingSuite}}\\
\cmidrule(lr){2-4} \cmidrule(lr){5-9}
 & \textbf{Pos.} & \textbf{Neg.} & \textbf{Avg.} & \textbf{Single} & \textbf{Stuff} & \textbf{Multi} & \textbf{Part} & \textbf{All} \\
\midrule
SAMTok-4B             & 43.5 & 80.4 & 61.9 & 80.9 & 12.4 & 62.0 & 52.9 & 57.5 \\
\;\;+Cap-Only GRPO      & 46.7 & \textbf{85.6} & 66.2 & 80.7 & 11.9 & 62.2 & 51.7 & 56.9 \\
\;\;+Loc-Only GRPO      & 43.2 & 80.4 & 61.8 & 90.7 & 18.4 & 67.7 & 60.3 & 65.0 \\
\;\;+CycleGRPO & \textbf{50.6} & 85.4 & \textbf{68.0} & \textbf{91.5} & \textbf{24.2} & \textbf{74.5} & \textbf{63.4} & \textbf{68.5} \\
\bottomrule
\end{tabular}}
\end{table}

\noindent\textbf{Comparison with GRPO baselines. }
To verify the necessity of the cycle-consistency reward, we compare CycleGRPO with various GRPO baselines. We first prompt the MLLM to generate captions for given regions. These generated captions are then used as text prompts for a subsequent referring segmentation task, where the resulting IoU serves as the reward signal for policy optimization.

The fundamental difference lies in the optimization target: while the standard GRPO baseline utilizes the spatial reconstruction accuracy (IoU) to tune the model, it lacks the direct, end-to-end incentive to optimize the semantic discriminability of the generated captions themselves. As shown in Table~\ref{tab:dlc_bench}, although standard GRPO achieves moderate gains on DLC-Bench (+1.5 Avg. score), it is significantly outperformed by CycleGRPO (+5.8 Avg. score). Furthermore, on the GRES benchmark (Table~\ref{tab:gres}), standard GRPO even suffers from a slight regression in segmentation precision (71.2 $\to$ 70.8 IoU). These results demonstrate that without the ``region $\to$ text $\to$ region'' loop, the model tends to fall into a sub-optimal policy that sacrifices descriptive richness for easier spatial matching.

We further compare with separate per-task (captioning-only or grounding-only) GRPO baselines, each of which optimizes a single direction and requires GT captions either as supervision or as input.
Our training data contains only (image, mask) pairs without caption GT, precluding direct per-task baselines on it. 
We thus sampled (image, mask, GT caption) triples from the DAM training set~\cite{dam} and trained both per-task baselines and CycleGRPO under matched compute, following Fig.~\ref{fig:teaser}. Captioning-only GRPO (image\,+\,mask $\to$ caption) uses an LLM-as-judge with Qwen2.5-72B on (predicted caption, GT caption) pairs; Localization-only GRPO (image\,+\,caption $\to$ mask) uses IoU as the reward. As shown in Tab.~\ref{tab:per_task_grpo}, each per-task baseline improves its own task but slightly drops the other. CycleGRPO, without using the GT caption, improves both and surpasses each baseline on its home metric, demonstrating mutual benefit from the cycle structure.

\noindent\textbf{Generalizability Across Grounding Formats. }
To investigate whether the efficacy of CycleGRPO is restricted to the SAMTok-based mask discretization, we conduct additional experiments using bounding boxes (bbox) as the spatial grounding format. In this setting, we utilize Qwen2.5VL-3B~\cite{qwen25vl} and Qwen3VL-4B\&8B~\cite{qwen3vl} as the base models, where region prompts for captioning and outputs for localization are represented in the standard bounding box format (e.g., \texttt{[y\_{min}, x\_{min}, y\_{max}, x\_{max}]}) instead of discrete mask tokens.

As shown in Table \ref{tab:bbox}, CycleGRPO consistently improves performance across vanilla MLLM backbones of varying scales, validating its generalizability beyond specific segmentation tokenizers like SAMTok. 
When applied to the bbox-based setting, CycleGRPO yields steady gains on DLC-Bench across all three backbones, lifting the average score by 2.9, 3.3, and 4.0 points on Qwen2.5-VL-3B, Qwen3-VL-4B, and Qwen3-VL-8B, respectively.
The benefits extend to spatial grounding as well: on GroundingSuite (Acc@0.5), CycleGRPO improves the overall score from 71.4 to 74.1 on both Qwen3-VL-4B and Qwen3-VL-8B. 
These results underscore that CycleGRPO is a generalizable framework that reinforces the spatial-semantic closed-loop across different model scales and grounding modalities, effectively scaling the model's fine-grained understanding regardless of the underlying spatial representation.

\noindent\textbf{Impact of Rollout Count $G$ and $K$.} 
We investigate the sensitivity of CycleGRPO to the hyperparameters $G$ and $K$. As shown in Table~\ref{tab:rollout}, increasing the sampling density generally leads to more stable feedback and better performance. Specifically, as $G$ and $K$ increase from 2 to 6, the average score on DLC-Bench rises from 61.9\% to 67.7\%, while the overall accuracy on GroundingSuite jumps from 57.5\% to 67.6\%. Notably, the performance on the challenging Part split nearly doubles, reaching 20.9\%. Although further scaling to $G=K=10$ still yields slight additional gains (e.g., 69.2\% on GroundingSuite), we select $G=K=6$ as our default configuration to maintain an optimal trade-off between computational cost and training efficacy. Additional ablation studies are provided in the supplementary materials.

\begin{table}[!tbp]
\centering\small
\caption{
Ablation on the number of rollouts $G$ and $K$.
We evaluate the impact of varying the captioning group size $G$ and localization rollout count $K$ on DLC-Bench and GroundingSuite. 
The results show that increasing the sampling density generally improves performance by providing more stable feedback. 
We select $G=K=6$ as our default configuration to maintain an optimal trade-off between training efficacy and computational cost.
}
\label{tab:rollout}
\setlength{\tabcolsep}{3pt}
\resizebox{0.75\columnwidth}{!}{%
\begin{tabular}{lccc|ccccc}
\toprule
\multirow{2}{*}{Rollout} & \multicolumn{3}{c|}{\textbf{DLC-Bench}} & \multicolumn{5}{c}{\textbf{GroundingSuite}} \\
\cmidrule(lr){2-4}
\cmidrule(lr){5-9}
& \textbf{Pos.} & \textbf{Neg.} & \textbf{Avg.} & \textbf{Stuff} & \textbf{Part} & \textbf{Multi} & \textbf{Single} & \textbf{All} \\
\midrule
G=K=0 (SAMTok) & 43.5 & 80.4 & 61.9 & 80.9 & 12.4 & 62.0 & 52.9 & 57.5\\
G=K=2   & 45.8 & 82.0 & 63.9 & 87.6 & 12.2 & 74.1 & 54.9 & 62.5 \\
G=K=6   & 51.2 & 84.2 & 67.7 & 90.7 & 20.9 & 76.3 & 61.6 & 67.6 \\
G=K=10  & 51.1 & 84.8 & 67.9 & 91.1 & 25.4 & 76.7 & 63.7 & 69.2 \\
\bottomrule
\end{tabular}}
\end{table}

\begin{table}[!tbp]
\centering
\caption{Ablation on spatial grounding formats on vanilla MLLM backbones.
We evaluate the generalizability of CycleGRPO by replacing the discrete mask tokens with continuous bounding box coordinates. 
}
\label{tab:bbox}
\resizebox{0.75\columnwidth}{!}{%
\begin{tabular}{lccc|ccccc}
\toprule
\multirow{2}{*}{\textbf{Method}} & \multicolumn{3}{c|}{\textbf{DLC-Bench}} & \multicolumn{5}{c}{\textbf{GroundingSuite (Acc@0.5)}} \\
\cmidrule(lr){2-4} \cmidrule(lr){5-9}
 & \textbf{Pos.} & \textbf{Neg.} & \textbf{Avg.} & \textbf{Single} & \textbf{Stuff} & \textbf{Multi} & \textbf{Part} & \textbf{All} \\
\midrule
Qwen2.5-VL-3B (bbox)      & 19.4 & 44.8 & 32.1 & 93.7 & 32.8 & 63.2 & 66.3 & 69.0 \\
\;\;+ CycleGRPO           & \textbf{23.3} & \textbf{46.6} & \textbf{35.0} & \textbf{94.1} & \textbf{34.1} & \textbf{64.1} & \textbf{67.3} & \textbf{69.9} \\
Qwen3-VL-4B (bbox)        & 39.6 & 62.4 & 51.0 & \textbf{95.2} & 24.2 & \textbf{76.7} & 67.0 & 71.4 \\
\;\;+ CycleGRPO           & \textbf{43.2} & \textbf{65.4} & \textbf{54.3} & 94.9 & \textbf{33.9} & 76.3 & \textbf{71.1} & \textbf{74.1} \\
Qwen3-VL-8B (bbox)        & 41.0 & \textbf{65.4} & 53.2 & 94.9 & 31.2 & 71.8 & 67.4 & 71.4 \\
\;\;+ CycleGRPO           & \textbf{49.4} & 65.0 & \textbf{57.2} & \textbf{95.9} & \textbf{37.6} & \textbf{74.4} & \textbf{70.4} & \textbf{74.1} \\
\bottomrule
\end{tabular}}
\end{table}

%% file: 2_related_work.tex
\section{Related Work}
\label{sec:related_work}

\noindent
\textbf{Region Understanding for MLLMs.}
Conventional MLLMs have demonstrated remarkable capabilities in image-level holistic understanding~\cite{qwen25vl, deepseekvl2, internvl35, wang2025ross3d, wang2025reconstructive, llava, o3, wang2025traceable, wang2025vgr, chen2025auto}.
To achieve a much more fine-grained visual comprehension, region-aware MLLMs have been proposed~\cite{dam, gar, osprey, guo2024regiongpt, ferret2, pixelrefer, videorefer, zhou2026samtok}.
However, most of these methods rely heavily on \textit{high-quality SFT data}, \textit{e.g.}, detailed region captions, which incur high annotation costs and are poorly scalable in real-world scenarios.
To address this, we present the \textit{actor as its own critic}, a scalable cycle reinforcement learning pipeline that requires only region inputs.
By leveraging cycle-consistent RL rewards, \textit{i.e.}, reconstructive IoU, the model is incentivized to generate discriminative descriptions of similar regions that capture unique details, thereby facilitating precise region understanding and localization.

\noindent\textbf{Visual Region Localization for MLLMs.}
Precise region localization is a core capability of modern MLLMs~\cite{qwen3vl,step3vl,internvl35,cogvlm2,glm45v}. 
Existing post-training approaches~\cite{wang2023visionllm,vectorllm,lisa,sa2va,zhang2024omg,padt,zhang2025pixel,himtok,alto,argenseg,lan2024text4seg,zhou2026samtok} typically enhance this ability by using instruction-based visual grounding dialogues that rely on large-scale, high-quality expression–location annotations. 
Such dependence creates a data bottleneck that limits further performance gains.
We address this issue with a CycleGRPO framework that requires only region locations as supervision. 


\noindent\textbf{RL for Region Understanding and Localization.}
In RL post-training for region understanding and localization, prior works~\cite{zhou2026samtok,caprl,wang2025traceable} typically optimize grounding and captioning separately. 
For visual grounding~\cite{univgr1,lens,segzero,segr1,sarch2025grounded,guig1,groundr1}, the rewards are calculated from distances or IoU between predicted locations and ground truth, which requires high-quality expression–location pairs. 
For region captioning~\cite{gar,dam,sun2026perceptiondlm}, reward design is more complex: methods such as CapRL~\cite{caprl} evaluate captions indirectly via question answering, relying on curated QA pairs and additional LLMs for reward computation, thereby increasing data and system costs.
In contrast, our framework unifies two tasks under a shared objective, simplifies the training pipeline, and reduces reliance on auxiliary data and models.

%% file: 5_conclusion.tex
\section{Conclusion}
\label{sec:conclusion}

This paper presented \textbf{CycleGRPO}, a reinforcement learning framework that leverages spatial-semantic duality to bootstrap multimodal capabilities without human-annotated region-caption pairs. 
By framing region understanding and localization as a self-evaluating closed loop, our method derives training signals from the cycle consistency between generated text and reconstructed spatial regions. 
Extensive experiments show that CycleGRPO significantly enhances both understanding and grounding across multiple benchmarks, outperforming much larger closed-source and open-source frontiers and generalizing to different spatial representations such as continuous bounding boxes. 
We argue that the ``Actor-as-Own-Critic'' paradigm offers a scalable path for the continued evolution of multimodal models without expensive human data.

%% file: 6_appendix.tex
\appendix

\noindent \textbf{Overview.} In this supplementary material, we present more detailed experimental results in addition to the main paper:
\begin{itemize}
    \item \textbf{Sec. \ref{sec:implementation_details}} provides details on the designed prompts and reward functions.
    \item \textbf{Sec. \ref{sec:gm}} extends our evaluation to the GroundingME benchmark to showcase fine-grained discrimination in ambiguous scenes.
    \item \textbf{Sec. \ref{sec:dlc_eval}} reports CycleGRPO under various DLC-Bench evaluation modes.
    \item \textbf{Sec. \ref{sec:cyclegrpo_8b}} provides an ablation study of CycleGRPO on the larger SAMTok-8B backbone.
    \item \textbf{Sec. \ref{sec:training_dynamics}} analyzes the training dynamics of CycleGRPO against single-task GRPO baselines.
    \item \textbf{Sec. \ref{sec:qual_bbox}} presents qualitative results under the bounding-box output mode.
    \item \textbf{Sec. \ref{sec:data_scaling}} provides an ablation study on the scaling properties of training data.
    \item \textbf{Sec. \ref{sec:ethics}} provides an analysis of ethics-related considerations and potential societal impacts.
    \item Finally, \textbf{Sec. \ref{sec:illustration}} offers qualitative visualizations of the spatial-semantic closed-loop in action.
\end{itemize}

\section{Implementation Details.}
\label{sec:implementation_details}

\noindent \textbf{Prompts.} To ensure consistent behavior across dual tasks, we design specific prompt templates for both region captioning and referring segmentation. Following the SAMTok architecture, each region mask is discretized and represented as a pair of special tokens. Specifically, a spatial region is encapsulated by the format $\mathcal{T}=$ \texttt{<|mt\_start|><|mt\_xxxx|><|mt\_xxxx|><|mt\_end|>}, where each \texttt{<|mt\_xxxx|>} denotes a unique codebook index corresponding to a spatial cluster.

\begin{itemize}
    \item \textbf{Region Captioning Prompt.} For the understanding task, the model is provided with the discretized mask tokens and tasked with generating a high-fidelity description. The canonical input prompt is constructed as follows:
    \begin{quote}
        \texttt{Provide a detailed description of this region \\ <|mt\_start|><|mt\_xx|><|mt\_xx|><|mt\_end|>. }
    \end{quote}
    \item \textbf{Referring Segmentation Prompt.} For the localization task, the model is required to map a semantic description back into the discrete token space. The input template is defined below, where the placeholder \texttt{\{description\}} is populated by the sampled region descriptions generated during the region captioning rollout phase:
    \begin{quote}
        \texttt{All spatial relationships are defined from the viewer's \\ perspective, where `front' means closer to the viewer and `back' means farther from the viewer. Please provide the \\ segmentation mask of the object the following statement \\ describes: \{description\}. Ensure that all details mentioned about the object are accurate. If a matching object is found, provide its segmentation mask in the format \\ `<|mt\_start|><|mt\_xxxx|><|mt\_xxxx|><|mt\_end|>'. If no matching object is found, output null.}
    \end{quote}
\end{itemize}

This structured prompt forces the model to focus on the fine-grained attributes mentioned in the caption, thereby facilitating cyclic reward computation in our \textit{CycleGRPO} framework.

\noindent \textbf{Reward Calculation.} 
To mitigate the substantial computational overhead and latency incurred by high-resolution mask decoding and pixel-level IoU computation during RL training, we propose a streamlined \textit{Hierarchical Token Grading} strategy. In the \textit{CycleGRPO} paradigm, the cyclic reward $R^{\rm total}$ is computed by comparing the reconstructed sequence $\hat{\mathcal{T}}$ (generated by the segmentor using the sampled \texttt{\{description\}}) against the original ground-truth sequence $\mathcal{T}$ that initiated the captioning rollout. Instead of projecting these tokens back into the image space, $R^{\rm total}$ is computed directly in the token domain using a heuristic-based grading function. Specifically, a full reward of $1.0$ is assigned for an exact string match between $\hat{\mathcal{T}}$ and $\mathcal{T}$. To provide denser feedback and facilitate reward shaping, we grant partial credits for near-miss predictions: a score of $0.8$ is awarded if the prefix including the first spatial token matches (i.e., identifying the correct primary spatial cluster), and $0.4$ is given if only the boundary anchors are correctly maintained with partial alignment. This hierarchical reward mechanism effectively converts a complex, dense prediction task into an efficient sequence-matching problem.

\section{Results on GroundingME Benchmark}
\label{sec:gm}
To further evaluate the generalization and fine-grained grounding capabilities of our approach, we conduct additional experiments on the GroundingME benchmark~\cite{li2025groundingme}.
GroundingME is specifically designed to assess models' ability to distinguish among multiple objects with only subtle differences within a single scene, requiring highly precise spatial-semantic associations.
We filter the dataset to exclude samples without valid target objects, yielding a curated evaluation set of 804 samples.
We utilize the provided ground-truth captions as input prompts to evaluate the segmentation performance (mIoU). This setup is particularly rigorous as the descriptions must resolve ambiguities among highly similar candidates to identify the correct target.
By incorporating CycleGRPO, the performance improves substantially from 29.5 (SAMTok) to 37.3.

\section{Various Evaluation Modes on DLC-Bench}
\label{sec:dlc_eval}
We strictly follow the official DLC-Bench evaluation pipeline according to GAR's~[29] code  (GitHub: {\small\texttt{EVALUATION.md}}). The main paper reports the ``Llama3.1-8B without images'' mode, where CycleGRPO surpasses even the SAMTok paper's caption-GT-finetuned variant. Tab.~\ref{tab:dlc_eval} additionally reports ``GPT-4o with images'' mode.

\begin{table}[!tbp]
\centering
\caption{Clarification on the DLC-Bench Evaluation.}
\label{tab:dlc_eval}
\resizebox{0.6\columnwidth}{!}{%
\begin{tabular}{llccc}
\toprule
& & \multicolumn{3}{c}{DLC-Bench} \\
\cmidrule(lr){3-5}
Eval mode & Method & Pos. & Neg. & Avg. \\
\midrule
\multirow{2}{*}{Llama3.1-8B w/o img} 
& SAMTok-4B           & 43.5 & 80.4 & 61.9 \\
& \;\;+CycleGRPO   & \textbf{51.2} & 84.2 & \textbf{67.7} \\
\midrule
\multirow{2}{*}{GPT-4o w/ img} 
& SAMTok-4B           & 59.0 & 81.0 & 70.0 \\
& \;\;+CycleGRPO   & \textbf{63.9} & \textbf{84.4} & \textbf{74.1} \\
\bottomrule
\end{tabular}}
\end{table}

\section{Ablation Study: CycleGRPO on SAMTok-8B}
\label{sec:cyclegrpo_8b}
We additionally evaluate CycleGRPO on a larger base model, SAMTok pretrained on Qwen3-VL-8B. As shown in Table~\ref{tab:cyclegrpo_samtok_8b}, CycleGRPO consistently improves over the SAMTok-8B baseline on both region captioning and grounding.

\begin{table}[!tbp]
\centering
\caption{CycleGRPO on SAMTok pretrained on Qwen3-VL-8B.}
\label{tab:cyclegrpo_samtok_8b}
\resizebox{0.8\columnwidth}{!}{%
\begin{tabular}{lccc|ccccc}
\toprule
 & \multicolumn{3}{c|}{DLC-Bench} & \multicolumn{5}{c}{GroundingSuite (gIoU)}\\
\cmidrule(lr){2-4} \cmidrule(lr){5-9}
Method & Pos. & Neg. & Avg. & Single & Stuff & Multi & Part & Avg. \\
\midrule
SAMTok-8B           & 47.3 & 80.0 & 63.7 & 88.6 & \textbf{15.9} & 65.9 & \textbf{58.7} & 63.1 \\
\;\;+ CycleGRPO & \textbf{51.6} & \textbf{84.6} & \textbf{68.1} & \textbf{90.1} & 15.2 & \textbf{77.7} & 55.2 & \textbf{64.5} \\
\bottomrule
\end{tabular}}
\end{table}

\section{Training Dynamics of CycleGRPO}
\label{sec:training_dynamics}
\begin{figure}[t]
  \centering
  \includegraphics[width=1\linewidth]{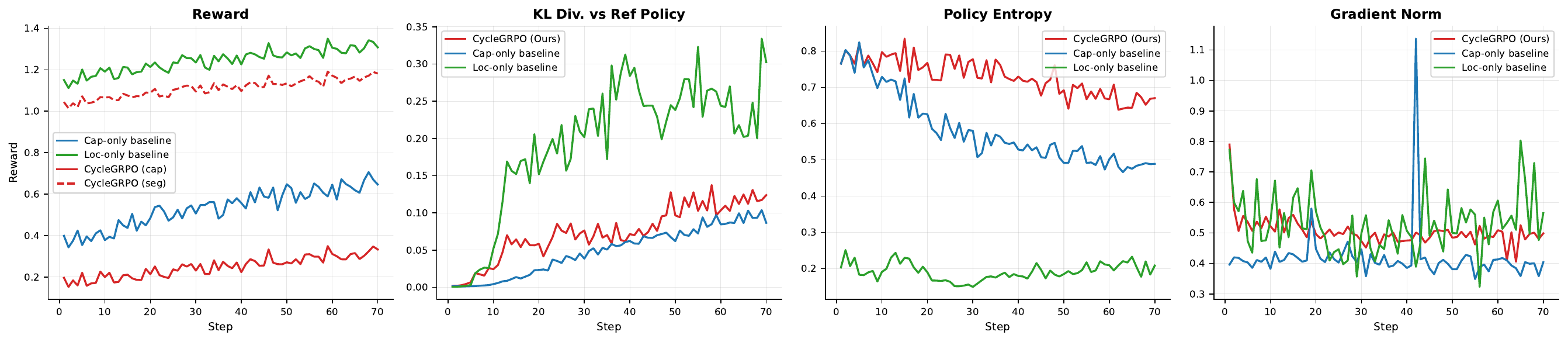}
  \caption{Training dynamics of CycleGRPO compared to the Cap-only and Loc-only GRPO baselines. From left to right: task rewards (CycleGRPO reports captioning and grounding rewards separately as solid and dashed lines), KL divergence from the reference policy, policy entropy, and gradient norm. CycleGRPO improves both reward components steadily while staying close to the reference policy and maintaining higher entropy than the Cap-only baseline, indicating stable training and sustained exploration.}
  \label{fig:training_dynamics}
\end{figure}

As Figure~\ref{fig:training_dynamics} shows, CycleGRPO trains stably and remains comparable to the single-task GRPO baselines, despite jointly optimizing captioning and grounding. The three runs correspond to the CycleGRPO, Cap-only GRPO, and Loc-only GRPO settings in Table 6 of the main paper. In the \emph{Reward} panel, both CycleGRPO reward components (solid: captioning, dashed: grounding) rise steadily, showing that neither sub-task stagnates. The \emph{KL Div.} panel shows the Loc-only baseline drifting far from the reference policy with large fluctuations, while CycleGRPO stays close to it, indicating that pairing the two tasks regularizes the policy. In \emph{Policy Entropy}, CycleGRPO maintains higher entropy than the steadily declining Cap-only baseline and the early-collapsing Loc-only baseline, preserving exploration and resisting mode collapse. Finally, all three runs keep a comparable, stable \emph{Gradient Norm}, confirming that the cycle-consistency objective adds no optimization instability.

\section{Qualitative Results under Bounding-Box Output}
\label{sec:qual_bbox}
\begin{figure}[t]
    \centering\small
    \includegraphics[width=0.8\linewidth]{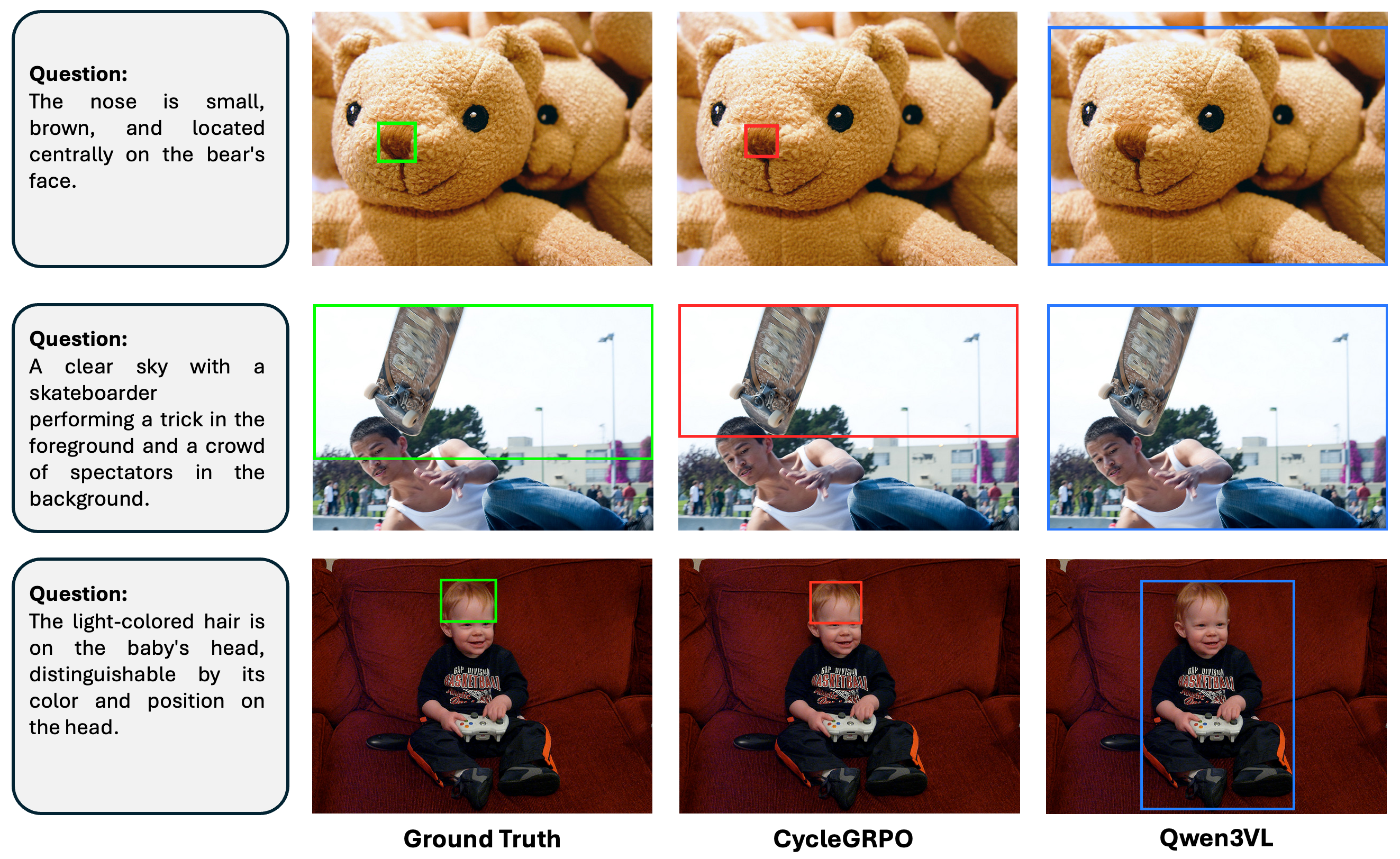}
    \caption{\textbf{Qualitative results on GroundingSuite under the bounding-box output mode.} We compare CycleGRPO (red) with the Qwen3VL baseline (blue) and the ground truth (green). Across part-level (the bear's nose, the baby's hair) and stuff-class (the clear sky) queries, SAMTok tends to predict overly large boxes covering the whole subject, while CycleGRPO produces tight boxes that are far better aligned with the queried region.}
    \label{fig:groundingsuite_bbox}
\end{figure}

In the main paper we visualize segmentation-mask predictions on GroundingSuite. Here we provide complementary results under the bounding-box output mode, where the model localizes the queried region with a single box. As shown in Figure~\ref{fig:groundingsuite_bbox}, the Qwen3VL-4B baseline tends to predict overly large boxes covering the whole subject, whereas CycleGRPO produces tight boxes closely matching the ground truth—across part-level (the bear's nose, the baby's hair) and stuff-class (the clear sky) queries alike. This indicates that the captioning–grounding cycle consistency sharpens spatial grounding for fine-grained textual queries.

\section{Ablation Study: Scaling of Training Data}
\label{sec:data_scaling}
\begin{figure}[H]
    \centering\small
    \includegraphics[width=0.46\linewidth]{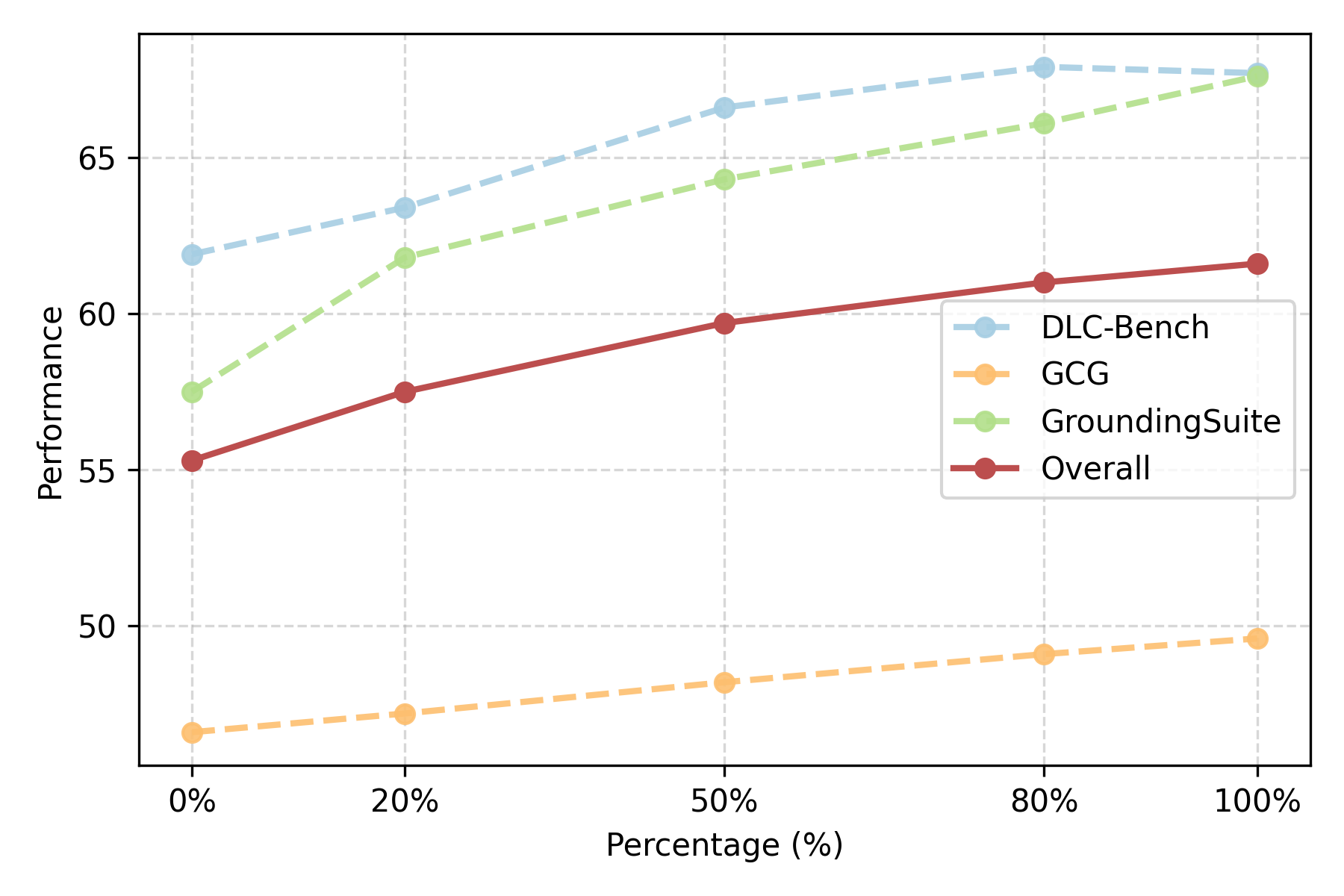}
    \caption{\small Ablation study on the scaling of training data.}
    \label{fig:scaling}
\end{figure}

To evaluate the data efficiency of CycleGRPO, we analyze the performance on DLC-Bench (Avg. Score), GCG (Recall on the val set), and GroundingSuite (All mIoU) across varying percentages of training data. As shown in Figure \ref{fig:scaling}, the average score exhibits a consistent upward trend. Notably, even with only 20\% or 50\% of the training data, the model achieves substantial gains, demonstrating the effectiveness of our self-evolutionary reward at harvesting meaningful gradients from fewer samples.

\section{Ethics Discussion}
\label{sec:ethics}
This work introduces CycleGRPO, a self-supervised reinforcement learning framework designed to enhance fine-grained understanding and localization in multimodal models. By using a cycle-consistency reward signal, the method reduces reliance on large-scale human annotations, thereby mitigating labor-intensive data collection issues. While the framework aims to reduce hallucinations and improve factual grounding, it remains subject to the inherent biases present in the pre-trained base models and the training data (e.g., DenseWorld, GRES). There is no intentional use of sensitive personal information or harmful content in the experiments; however, users should exercise caution when deploying the model in sensitive real-world applications to ensure that the generated captions or segmentations do not propagate societal biases or provide misleading information in high-stakes contexts.

\section{Illustrations of the Closed-loop in CycleGRPO}
\label{sec:illustration}
To provide a more intuitive understanding of how CycleGRPO fosters self-evolution without human intervention, we visualize the fine-grained generative process in Figure \ref{fig:rollout}. 
As conceptualized in our ``Actor-as-its-own-Critic'' paradigm, each training step forms a complete spatial-semantic closed-loop. 

Specifically, for a given target region, the model first performs a \textit{Captioning Rollout} (with $G=4$) to generate multiple descriptive hypotheses. These candidates are then immediately fed back into the model for a \textit{Localization Rollout} (with $K=6$) to reconstruct the original region solely from the generated text. 

The visualizations reveal a compelling correlation: captions that are semantically rich and factually grounded (highlighted in \textcolor{green}{green}) consistently yield higher IoU scores during the subsequent localization phase. 
Conversely, descriptions containing hallucinations or vague terminology (highlighted in \textcolor{red}{red}) lead to reconstruction failures. 
This discrepancy provides a naturally occurring, high-variance reward signal. By maximizing this cycle-consistency reward, CycleGRPO effectively compels the model to bridge its understanding and grounding capabilities, ensuring that every bit of semantic detail is spatially verifiable. 
This self-supervised mechanism allows the model to scale its fine-grained intelligence by exploring and rectifying its own internal logic, bypassing the need for expensive, static human-labeled pairs.

\begin{center}
    \centering\small
    \includegraphics[width=1.0\linewidth]{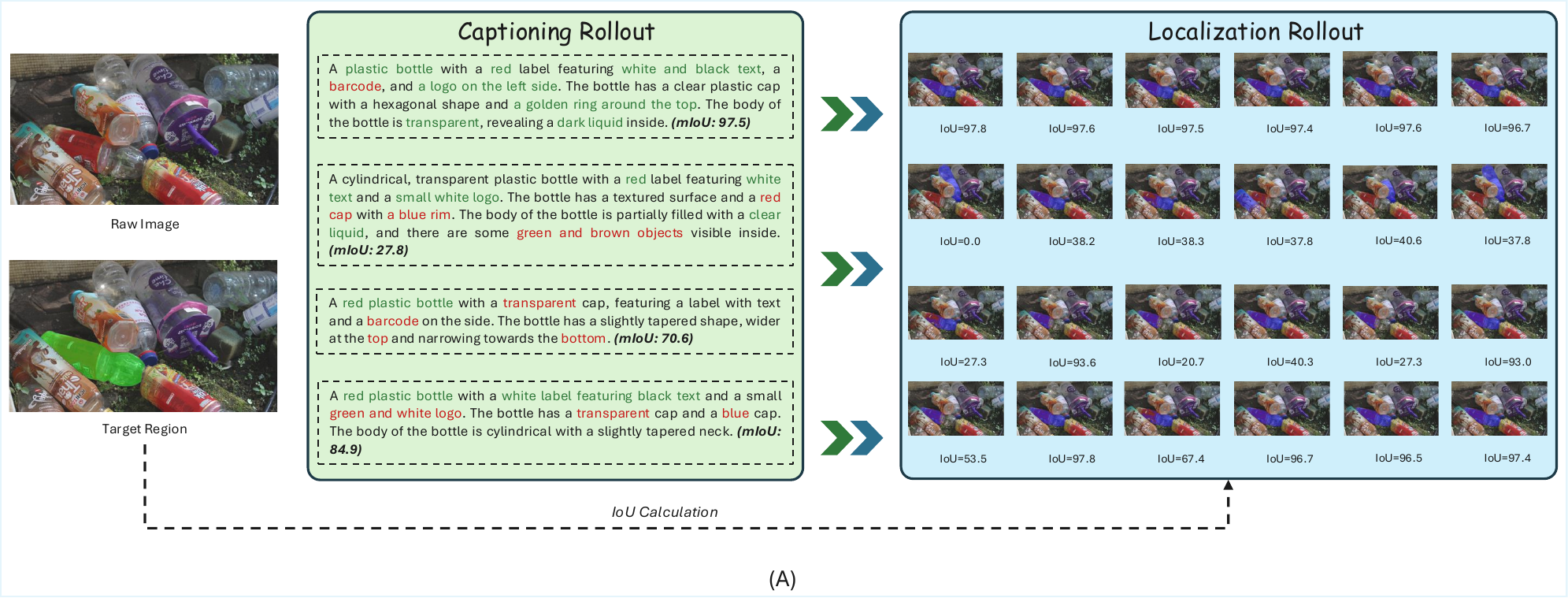}
    \includegraphics[width=1.0\linewidth]{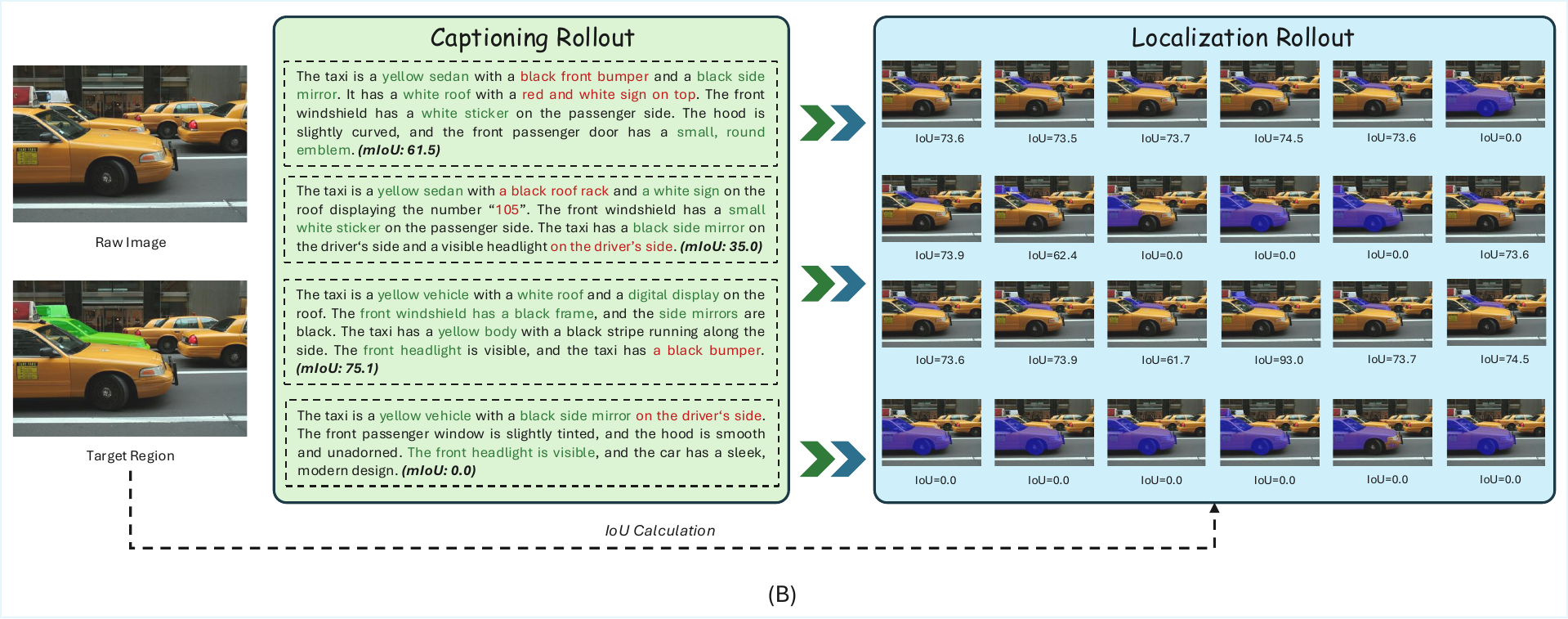}
    \includegraphics[width=1.0\linewidth]{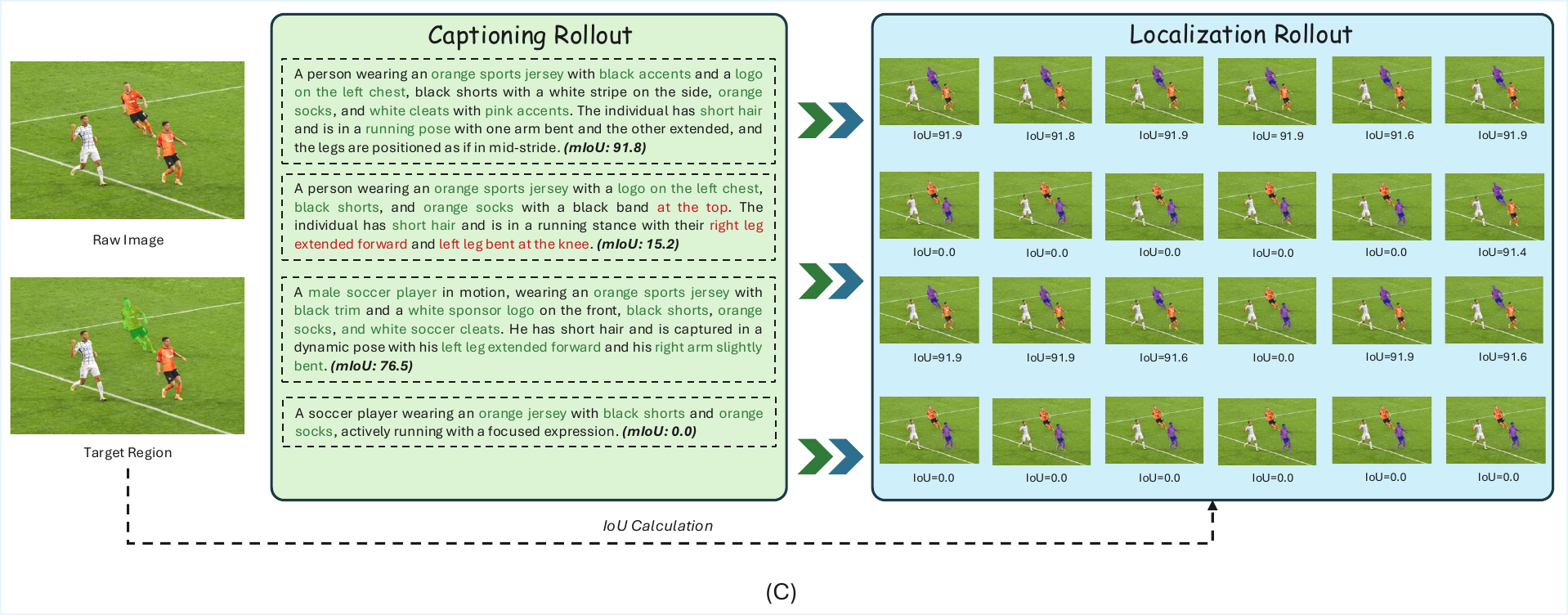}
    \includegraphics[width=1.0\linewidth]{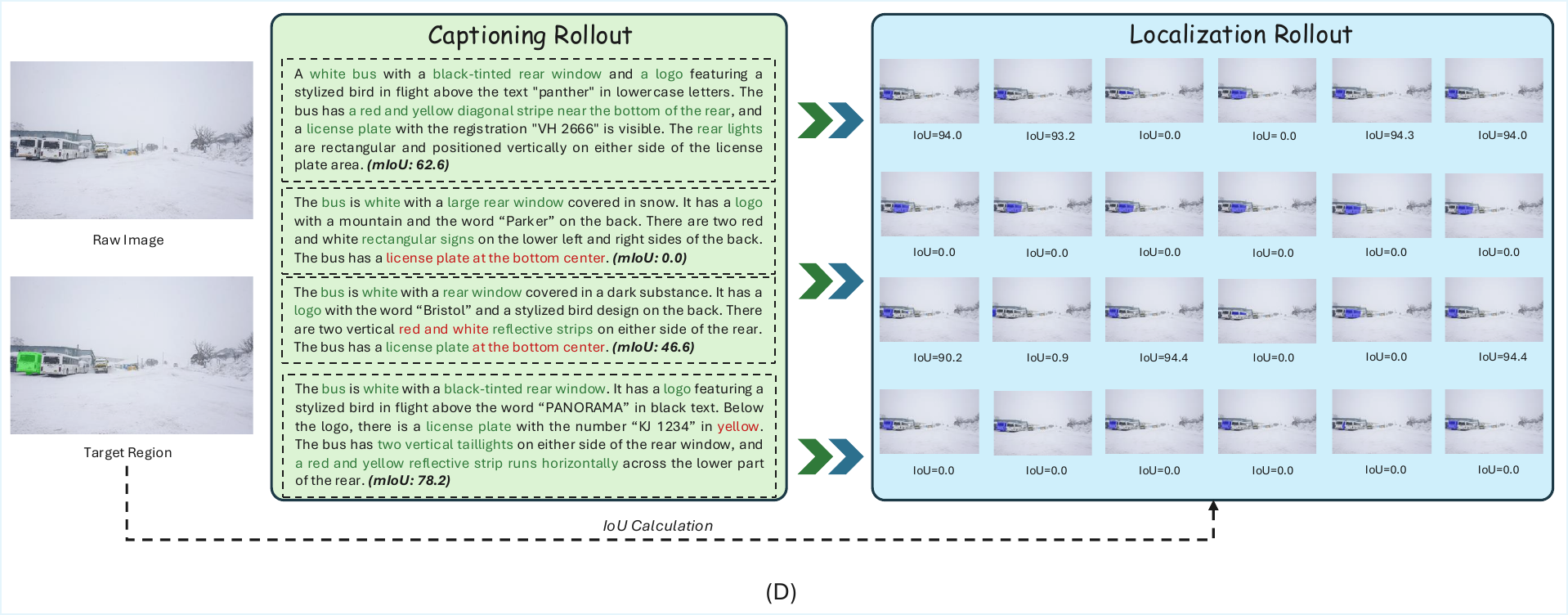}
    \includegraphics[width=1.0\linewidth]{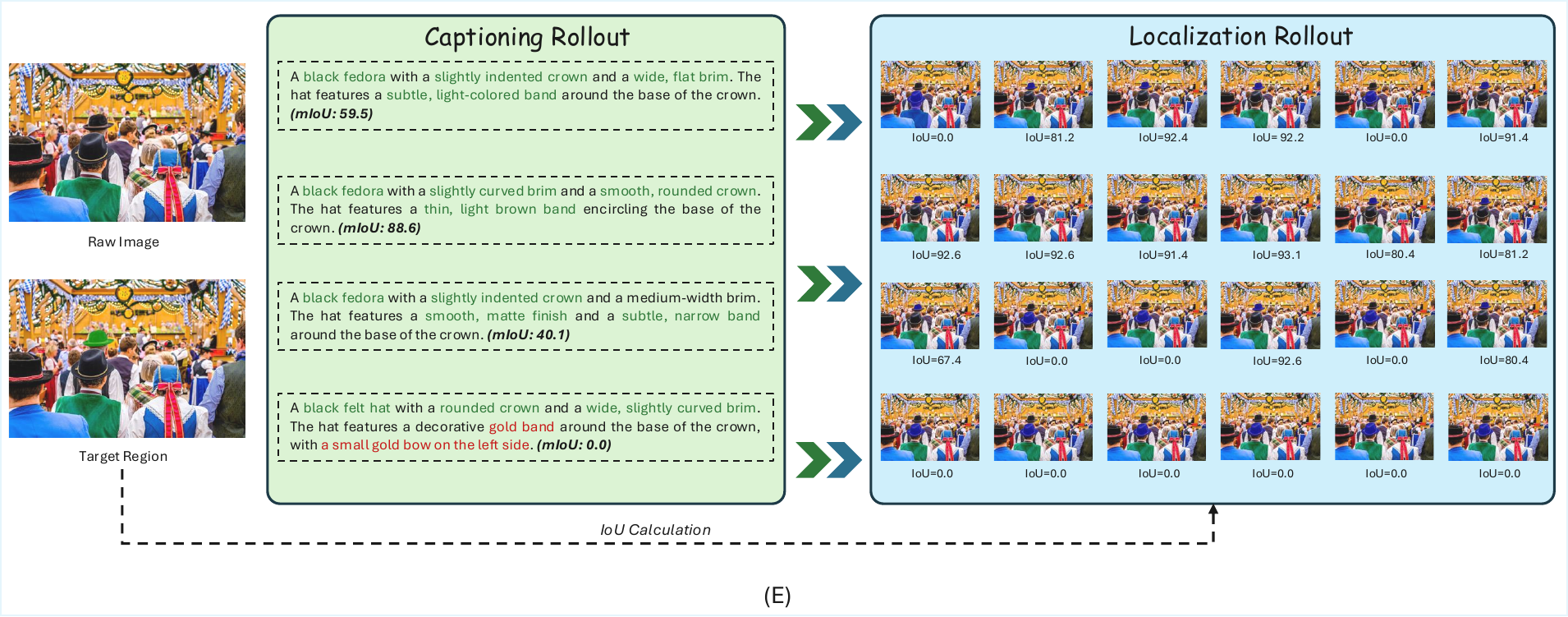}
    \includegraphics[width=1.0\linewidth]{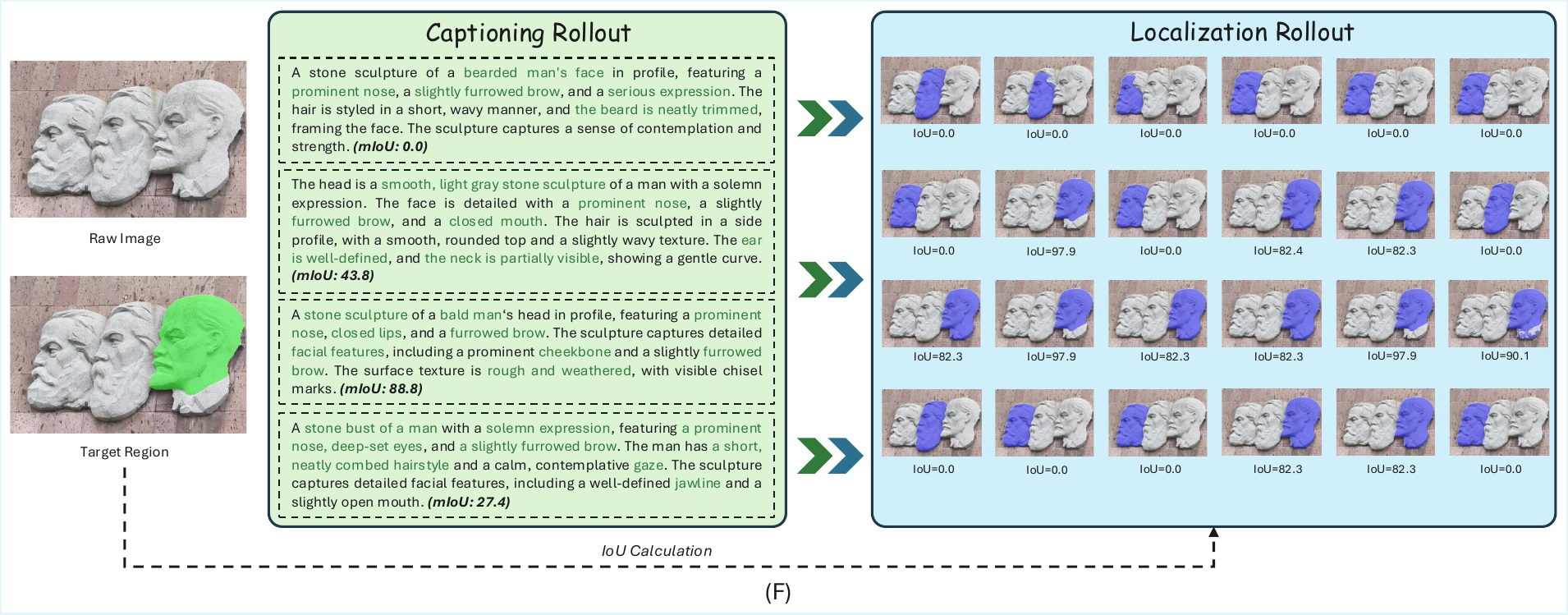}
    \includegraphics[width=1.0\linewidth]{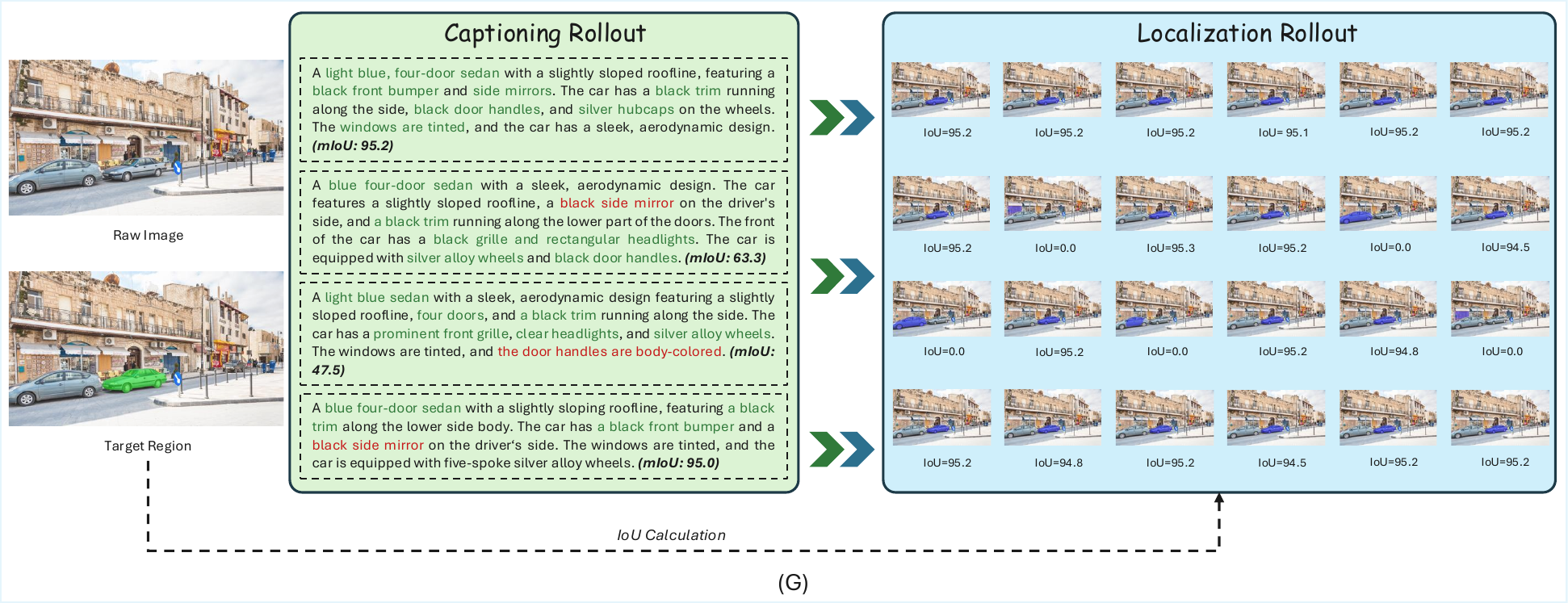}
    \includegraphics[width=1.0\linewidth]{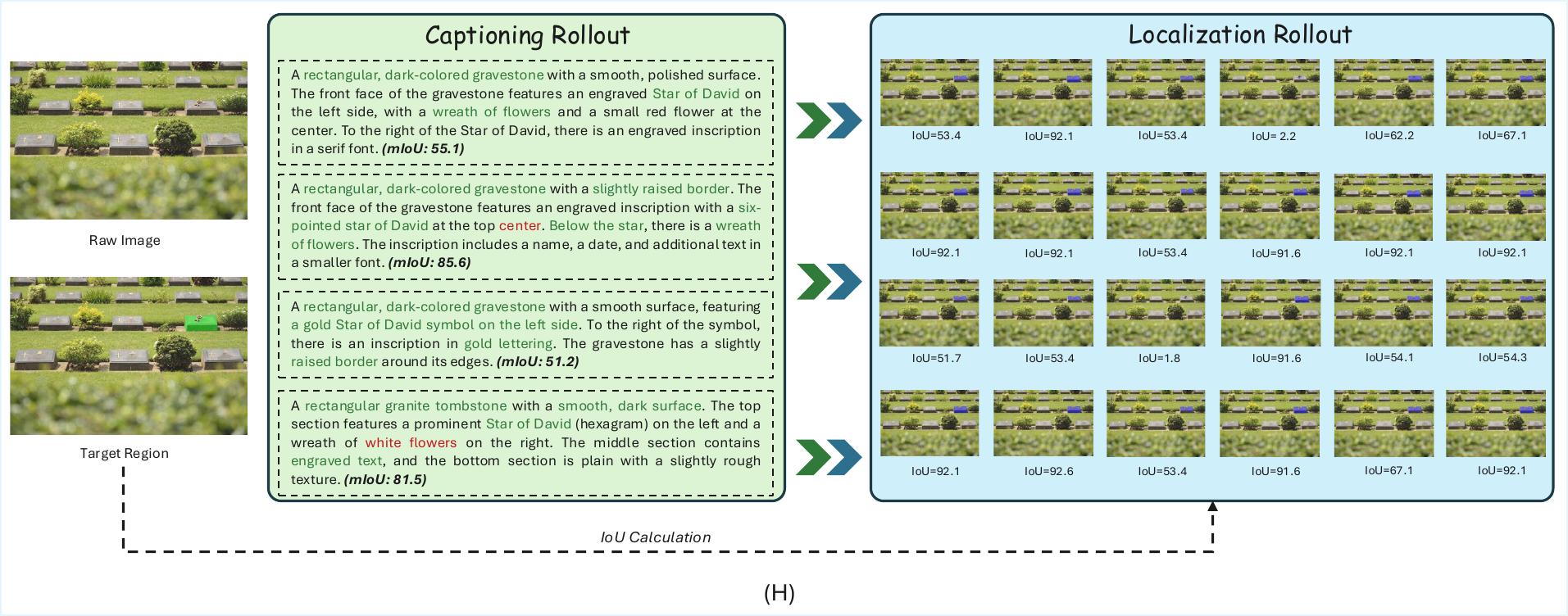}
    \includegraphics[width=1.0\linewidth]{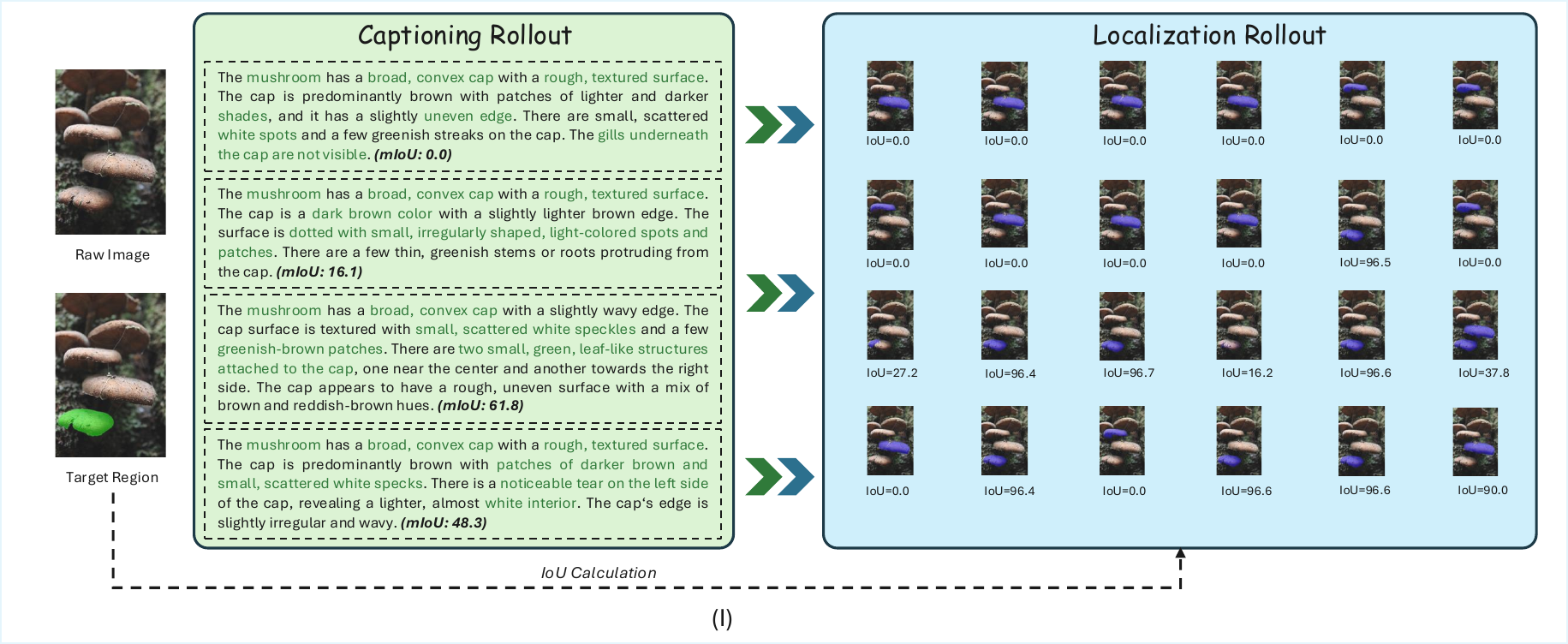}
    \captionof{figure}{\textbf{Visualization of the self-evolution rollouts under the CycleGRPO framework.} 
    We illustrate the generative process with $G=4$ (number of captioning rollouts) and $K=6$ (number of localization rollouts). 
    In the Target Region, the \textcolor{green}{green} mask denotes the ground-truth (GT) region. 
    In the \textit{Captioning Rollout}, phrases highlighted in \textcolor{green}{green} represent accurate semantic descriptions, while those in \textcolor{red}{red} indicate hallucinations or errors. 
    In the \textit{Localization Rollout}, the \textcolor{blue}{blue} masks represent the predicted regions based on the generated captions. 
    Consistent with the principle of \textbf{spatial-semantic self-consistency}, we observe that high mIoU scores are strongly correlated with both descriptive accuracy and precise localization. 
    Furthermore, captions containing more \textbf{unique and fine-grained key information} (e.g., object colors, logo positions) consistently yield superior localization performance compared to vague or partially incorrect descriptions.}
    \label{fig:rollout}
\end{center}